\documentclass{article}
\usepackage{microtype}
\usepackage{graphicx}
\usepackage{subcaption}  
\usepackage{booktabs} 
\usepackage{multirow, boldline}
\usepackage{tikz}
\usetikzlibrary{shapes.misc, positioning}

\usepackage{url}

\usepackage{breakurl}
\usepackage[breaklinks]{hyperref}

\usepackage{hyperref}

\usepackage[accepted]{icml2019}

\icmltitlerunning{Flexibly Fair Representation Learning by Disentanglement}

\usepackage{amssymb}
\usepackage{amsmath}
\usepackage{natbib}
\usepackage{xcolor}

\newcommand{\E}{\mathbb{E}}
\newcommand{\R}{\mathbb{R}}
\newcommand{\argmax}{\operatorname{argmax}}

\newif\ifcomments
\commentstrue
\ifcomments
  \newcommand{\colornote}[3]{{\color{#1}\bf{#2: #3}\normalfont}}
\else
  \newcommand{\colornote}[3]{}
\fi

\begin{document}

\twocolumn[
\icmltitle{Flexibly Fair Representation Learning by Disentanglement}



\icmlsetsymbol{equal}{*}

\begin{icmlauthorlist}
\icmlauthor{Elliot Creager}{to,vi}
\icmlauthor{David Madras}{to,vi}
\icmlauthor{J\"orn-Henrik Jacobsen}{vi}
\icmlauthor{Marissa A. Weis}{vi,tu}
\icmlauthor{Kevin Swersky}{go}
\icmlauthor{Toniann Pitassi}{to,vi}
\icmlauthor{Richard Zemel}{to,vi}
\end{icmlauthorlist}

\icmlaffiliation{to}{University of Toronto}
\icmlaffiliation{vi}{Vector Institute}
\icmlaffiliation{go}{Google Research}
\icmlaffiliation{tu}{University of T\"ubingen}

\icmlcorrespondingauthor{Elliot Creager}{creager@cs.toronto.edu}

\icmlkeywords{Machine Learning, ICML}

\vskip 0.3in
]



\printAffiliationsAndNotice{}  

\begin{abstract}
We consider the problem of learning representations that achieve group and subgroup fairness with respect to multiple sensitive attributes.
Taking inspiration from the disentangled representation learning literature, we propose an algorithm for learning compact representations of datasets that are useful for reconstruction and prediction, but are also \emph{flexibly fair}, meaning they can be easily modified at test time to achieve subgroup demographic parity with respect to multiple sensitive attributes and their conjunctions.
We show empirically that the resulting encoder---which does not require the sensitive attributes for inference---enables the adaptation of a single representation to a variety of fair classification tasks with new target labels and subgroup definitions.
\end{abstract}

\newcommand{\arrowWidth}{1.7pt}
\newcommand{\rectWidth}{0.5cm}
\newcommand{\circleSize}{0.9cm}
\begin{figure*}[htp]
\begin{center}
\begin{subfigure}[t]{0.45\textwidth}
\begin{tikzpicture}
\path (0.0,0.0) node[rectangle,draw,minimum size=\circleSize,align=center](x) {
    $x$ \includegraphics[scale=0.25]{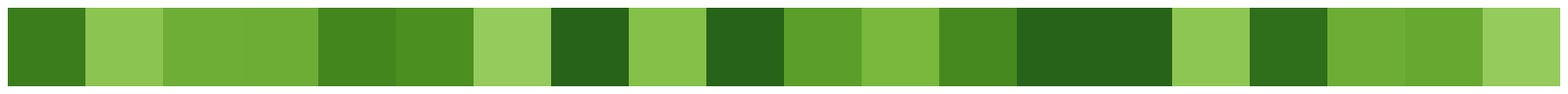}\\\small non-sensitive observations
}
(4.2,0.0) node[rectangle,draw,minimum size=\circleSize,align=center](a) {
    $a$ \includegraphics[scale=0.08]{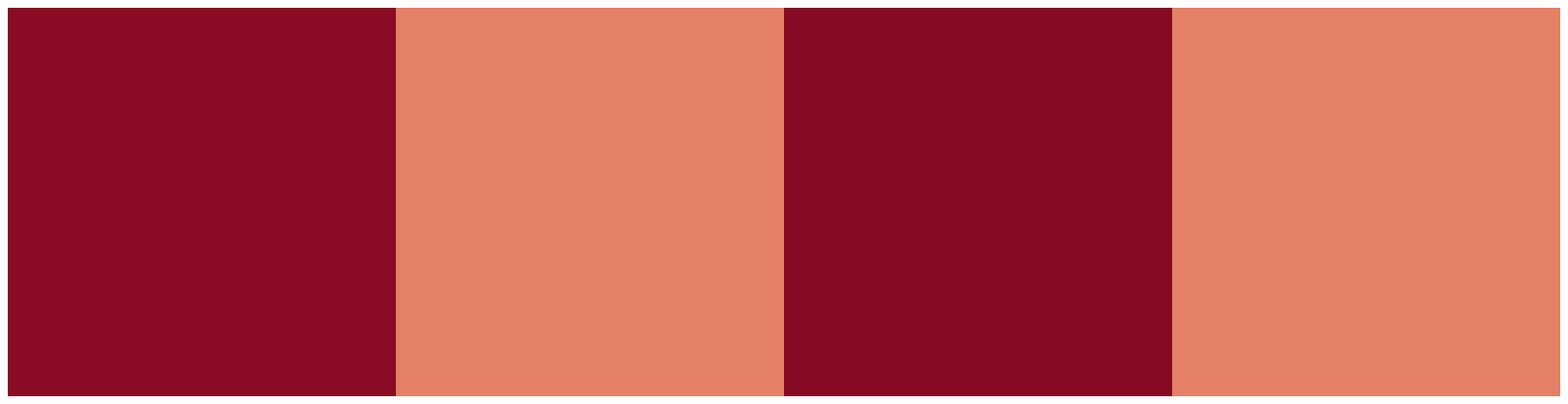}\\\small sensitive observations
}
(0.0,3.0) node[rectangle,draw,minimum size=\circleSize,align=center](z) {
    $z$ \includegraphics[scale=0.2]{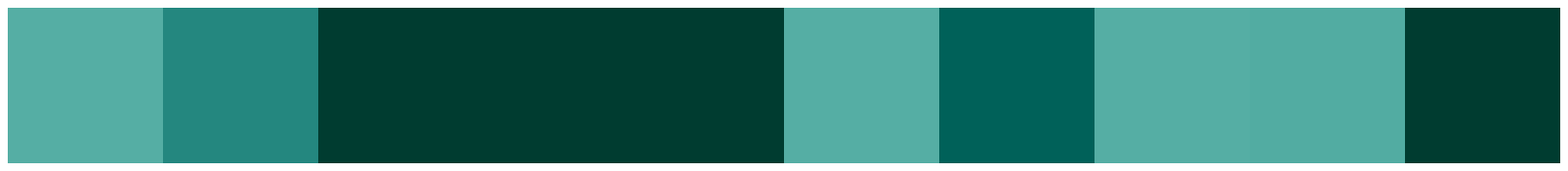}\\\small non-sensitive latents
}
(4.2,3.0) node[rectangle,draw,minimum size=\circleSize,align=center](b) {
    $b$ \includegraphics[scale=0.08]{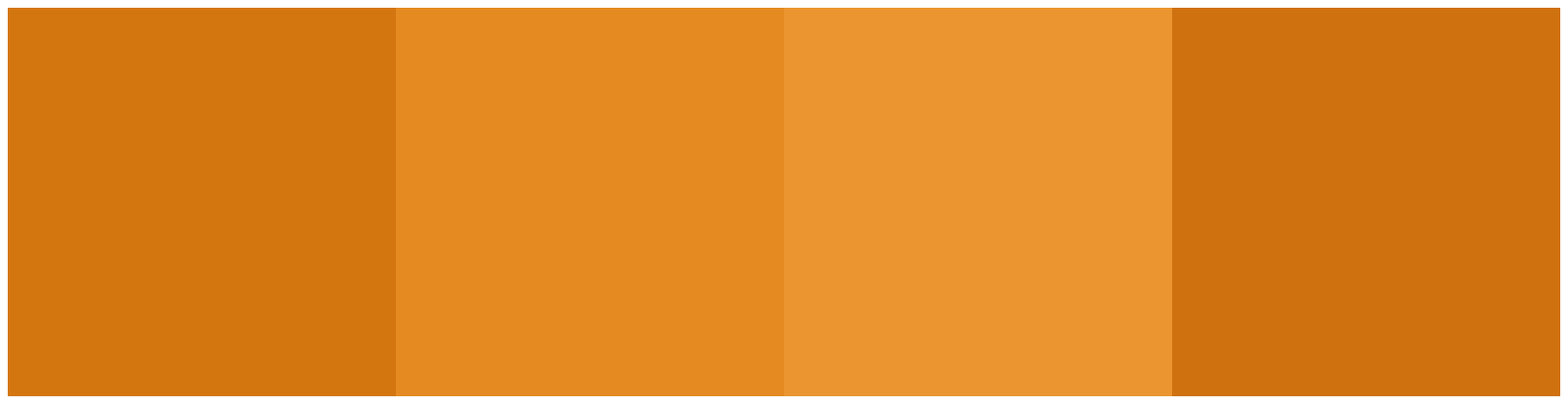}\\\small sensitive latents
};
\draw[->,>=stealth, line width=\arrowWidth] (x) to [out=45,in=-45,looseness=1.0] (z);
\draw[->,>=stealth, line width=\arrowWidth] (z) to [out=-135,in=135,looseness=1.0] (x);
\draw[->,>=stealth, line width=\arrowWidth] (x) to [out=20,in=-110,looseness=1.0] (b);
\draw[->,>=stealth, line width=\arrowWidth] (b) to [out=-130,in=40,looseness=1.0] (x);
\draw[->,>=stealth, line width=\arrowWidth] (b) -- (a);
\end{tikzpicture}
    \caption{
    FFVAE learns the encoder distribution $q(z,b|x)$ and decoder distributions $p(x|z,b)$, $p(a|b)$ from inputs $x$ and multiple sensitive attributes $a$.
    The disentanglement prior structures the latent space by encouraging low $\text{MI}(b_i,a_j)\forall i \neq j$ and low $\text{MI}(b,z)$ where $\text{MI}(\cdot)$ denotes mutual information.
    }
\label{fig:1a}
\end{subfigure}
\hfill
\begin{subfigure}[t]{0.45\textwidth}
\begin{tikzpicture}
\path
(0.0-1.8,0.0) node[rectangle,draw,minimum size=\circleSize,align=center](x) {
    $x$ \includegraphics[scale=0.25]{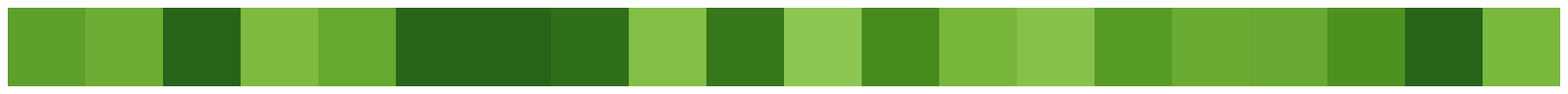}
}
(0.0-1.8,2.2) node[rectangle,draw,minimum size=\circleSize,align=center](z) {
    $z$ \includegraphics[scale=0.2]{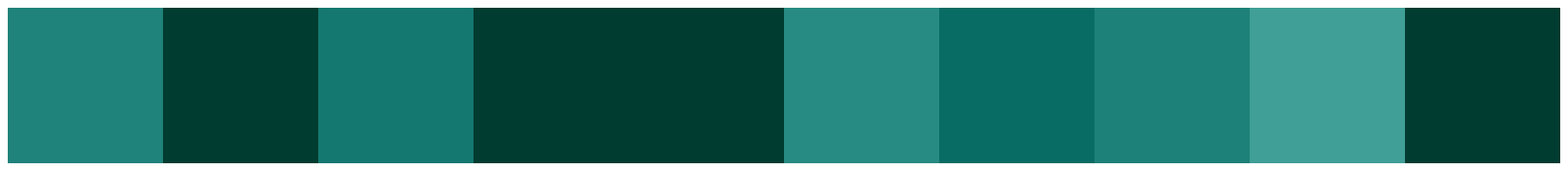}
}
(4.0-1.8,1.1) node[rectangle,draw,minimum size=\circleSize,align=center](b) {
    $b$ \includegraphics[scale=0.08]{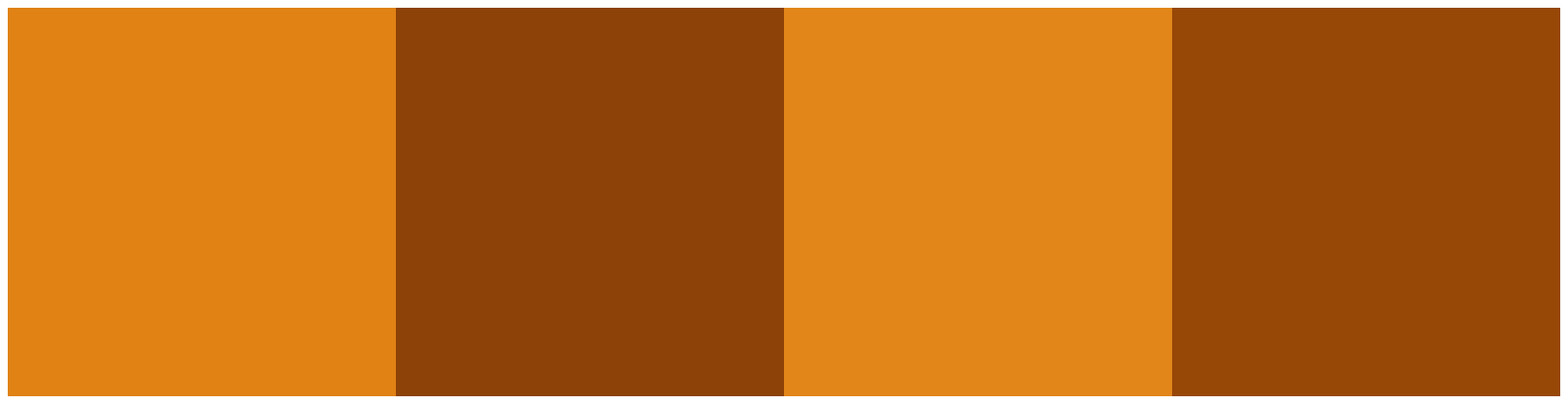}
}
(4.0-1.8,2.5) node[rectangle,draw,minimum size=\circleSize,align=center](bprime) {
    $b'$ \includegraphics[scale=0.08]{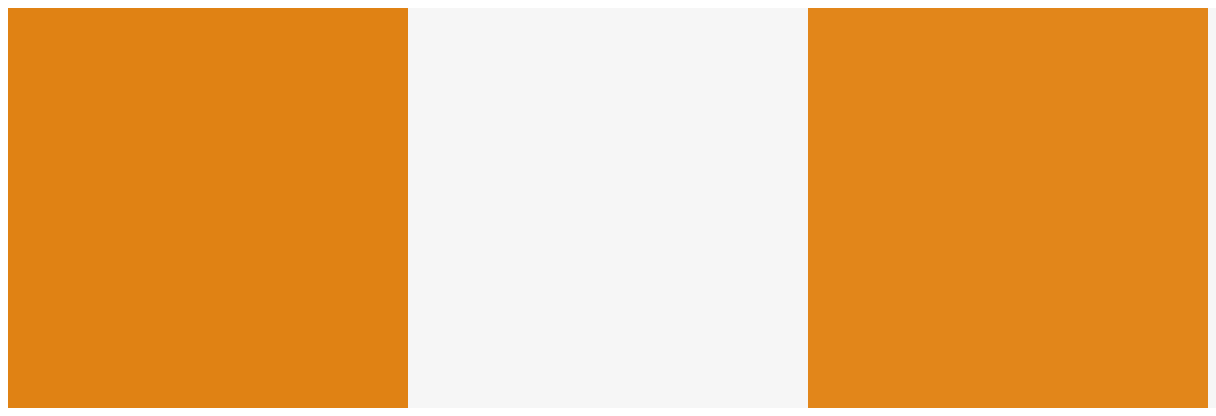}\\\scriptsize modified sens. latents
}
(2.0-1.8,4.0) node[rectangle,draw,minimum size=\circleSize,align=center](y) {
    $y$ \includegraphics[scale=0.2]{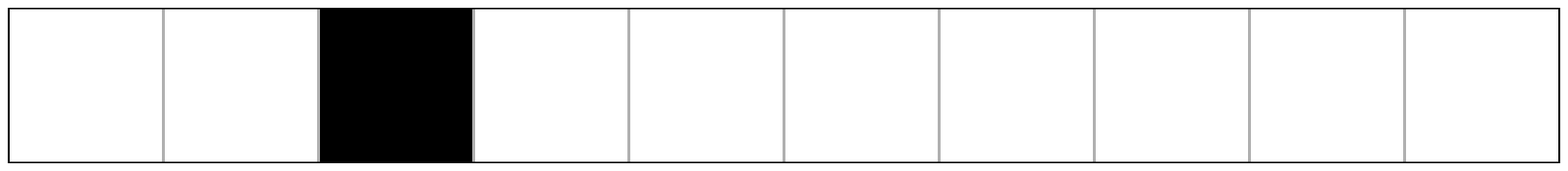}\\\small target label
};
\draw[->,>=stealth, line width=\arrowWidth] (x) -- (z);
\draw[->,>=stealth, line width=\arrowWidth] (x) to [out=0,in=-90,looseness=1.6] (b);
\draw[->,>=stealth, line width=\arrowWidth] (b) -- (bprime);
\draw[->,>=stealth, line width=\arrowWidth] (bprime) -- (y);
\draw[->,>=stealth, line width=\arrowWidth] (z) -- (y);
\end{tikzpicture}
    \caption{The FFVAE latent code $[z,b]$ can be modified by discarding or noising out sensitive dimensions $\{b_j\}$, which yields a latent code $[z, b']$ independent of groups and subgroups derived from sensitive attributes $\{a_j\}$.
A held out label $y$ can then be predicted with subgroup demographic parity.
}

\label{fig:1b}
\end{subfigure}
\caption{
    Data flow at train time (\ref{fig:1a}) and test time (\ref{fig:1b}) for our model, Flexibly Fair VAE (FFVAE).
}
\label{modelpic}
\end{center}
\vskip -0.2in
\label{fig:1}
\end{figure*}

\section{Introduction}
Machine learning systems are capable of exhibiting discriminatory behaviors against certain demographic groups in high-stakes domains such as law, finance, and medicine \citep{kirchner2016machine,aleo2008foreclosure,kim2015sex}.
These outcomes are potentially unethical or illegal \citep{barocas2014datas,hellman2017indirect}, and behoove researchers to investigate more equitable and robust models.
One promising approach is fair representation learning: the design of neural networks using learning objectives that satisfy certain fairness or parity constraints in their outputs \citep{zemel2013learning,louizos2015variational,edwards2015censoring,madras2018learning}.
This is attractive because neural network representations often generalize to tasks that are unspecified at train time, which implies that a properly specified fair network can act as a group parity bottleneck that reduces discrimination in unknown downstream tasks.

Current approaches to fair representation learning are flexible with respect to downstream tasks but inflexible with respect to sensitive attributes.
While a single learned representation can adapt to the prediction of different task labels~$y$, the single sensitive attribute $a$ for \emph{all} tasks must be specified at train time.
Mis-specified or overly constraining train-time sensitive attributes could negatively affect performance on downstream prediction tasks.
Can we instead learn a \textit{flexibly fair} representation that can be \emph{adapted}, at test time, to be fair to a variety of protected groups and their intersections?
Such a representation should satisfy two criteria.
Firstly, the structure of the latent code should facilitate \textit{simple} adaptation, allowing a practitioner to easily adapt the representation to a variety of fair classification settings, where each task may have a different task label $y$ and sensitive attributes $a$.
Secondly, the adaptations should be \textit{compositional}: the representations can be made fair with respect to conjunctions of sensitive attributes, to guard against subgroup discrimination (e.g., a classifier that is fair to women but not Black women over the age of 60).
This type of subgroup discrimination has been observed in commercial machine learning systems \citep{buolamwini2018gender}.

In this work, we investigate how to learn flexibly fair representations that can be easily adapted at test time to achieve fairness with respect to sets of sensitive groups or subgroups.
We draw inspiration from the disentangled representation literature, where the goal is for each dimension of the representation (also called the ``latent code'') to correspond to no more than one semantic factor of variation in the data (for example, independent visual features like object shape and position) \citep{higgins2016beta, locatello2018challenging}.
Our method uses multiple sensitive attribute labels at train time to induce a disentangled structure in the learned representation, which allows us to easily eliminate their influence at test time.
Importantly, at test time our method does not require access to the sensitive attributes, which can be difficult to collect in practice due to legal restrictions \citep{elliot2008new,division1983}.
The trained representation permits simple and composable modifications at test time that eliminate the influence of sensitive attributes, enabling a wide variety of downstream tasks.

We first provide proof-of-concept by generating a variant of the synthetic DSprites dataset with correlated ground truth factors of variation, which is better suited to fairness questions.
We demonstrate that even in the correlated setting, our method is capable of disentangling the effect of several sensitive attributes from data, and that this disentanglement is useful for fair classification tasks downstream.
We then apply our method to a real-world tabular dataset (Communities \& Crime) and an image dataset (Celeb-A), where we find that our method matches or exceeds the fairness-accuracy tradeoff of existing disentangled representation learning approaches on a majority of the evaluated subgroups.

\section{Background} \label{sec:background}
\paragraph{Group Fair Classification}
In fair classification, we consider labeled examples $x, a, y \sim p_{\text{data}}$ where $y \in \mathcal{Y}$ are labels we wish to predict, $a \in \mathcal{A}$ are \emph{sensitive} attributes, and $x \in \mathcal{X}$ are non-sensitive attributes. 
The goal is to learn a classifier $\hat y = g(x, a)$ (or $\hat y = g(x)$) which is predictive of $y$ and achieves certain group fairness criteria w.r.t. $a$.
These criteria are typically written as independence properties of the various random variables involved.
In this paper we focus on demographic parity, which is satisfied when the predictions are independent of the sensitive attributes: $\hat y \perp a$.
It is often impossible or undesirable to satisfy demographic parity exactly (i.e. achieve complete independence).
In this case, a useful metric is \textit{demographic parity distance}:
\begin{equation}
    \Delta_{DP} = | \E [\bar y = 1 | a = 1] - \E [\bar y = 1 | a = 0] |
\end{equation}
where $\bar y$ here is a binary prediction derived from model output $\hat y$.
When $\Delta_{DP}=0$, demographic parity is achieved; in general, lower $\Delta_{DP}$ implies less unfairness.

Our work differs from the fair classification setup as follows:
We consider several sensitive attributes at once, and seek fair outcomes with respect to each; individually as well as jointly (cf. subgroup fair classification, \citet{kearns2017preventing,johnson2018multicalibration});
also, we focus on representation learning rather than classification, with the aim of enabling a range of fair classification tasks downstream.

\paragraph{Fair Representation Learning}
In order to flexibly deal with many label and sensitive attribute sets, we employ representation learning to compute a compact but predicatively useful encoding of the dataset that can be flexibly adapted to different fair classification tasks.
As an example, if we learn the function $f$ that achieves independence in the representations $z \perp a$ with $z=f(x, a)$ or $z=f(x)$, then any predictor derived from this representation will also achieve the desired demographic parity, $\hat y \perp a$ with $\hat y = g(z)$.

The fairness literature typically considers binary labels and sensitive attributes: $\mathcal{A} = \mathcal{Y} = \{0, 1\}$.
In this case, approaches like regularization \citep{zemel2013learning} and adversarial regularization \citep{edwards2015censoring, madras2018learning} are straightforward to implement. 
We want to address the case where $a$ is a vector with many dimensions.
Group fairness must be achieved for each of the dimensions in $a$ (age, race, gender, etc.) and their combinations.

\paragraph{VAE}
The vanilla Variational Autoencoder (VAE) \citep{kingma2013auto} is typically implemented with an isotropic Gaussian prior $p(z) = \mathcal{N}(0, I)$.
The objective to be maximized is the Evidence Lower Bound (a.k.a., the ELBO), 
\begin{align}
    L_{\text{VAE}}(p,q) = \E_{q(z|x)} \left[ \log p(x|z) \right] - D_{KL} \left[ q(z|x) || p(z)  \right], \nonumber
\end{align}
which bounds the data log likelihood $\log p(x)$ from below for any choice of $q$.
The encoder and decoder are often implemented as Gaussians 
\begin{align}
    q(z|x) &= \mathcal{N}(z|\mu_q(x),  \Sigma_q(x)) \nonumber \\
    p(x|z) &= \mathcal{N}(x|\mu_p(z), \Sigma_p(z)) \nonumber
\end{align}
whose distributional parameters are the outputs of neural networks $\mu_q(\cdot)$, $\Sigma_q(\cdot)$, $\mu_p(\cdot)$, $\Sigma_p(\cdot)$, with the $\Sigma$ typically exhibiting diagonal structure.
For modeling binary-valued pixels, a Bernoulli decoder $p(x|z)=\text{Bernolli}(x|\theta_p(z))$ can be used.
The goal is to maximize $L_{\text{VAE}}$---which is made differentiable by reparameterizing samples from $q(z|x)$---w.r.t. the network parameters.

\paragraph{$\beta$-VAE}
\citet{higgins2016beta} modify the VAE objective:
\begin{align}
    L_{\beta\text{VAE}}(p,q) = \E_{q(z|x)} \left[ \log p(x|z) \right] - \beta D_{KL} \left[ q(z|x) || p(z) \right ]. \nonumber
\end{align}
The hyperparameter $\beta$ allows the practitioner to encourage the variational distribution $q(z|x)$ to reduce its KL-divergence to the isotropic Gaussian prior $p(z)$.
With $\beta>1$ this objective is a valid lower bound on the data likelihood. 
This gives greater control over the model's adherence to the prior.
Because the prior factorizes per dimension $p(z) = \prod_j p(z_j)$, \citet{higgins2016beta} argue that increasing $\beta$ yields ``disentangled'' latent codes in the encoder distribution $q(z|x)$.
Broadly speaking, each dimension of a properly disentangled latent code should capture no more than one semantically meaningful factor of variation in the data.
This allows the factors to be manipulated in isolation by altering the per-dimension values of the latent code.
Disentangled autoencoders are often evaluated by their sample quality in the data domain, but we instead emphasize the role of the encoder as a representation learner to be evaluated on downstream fair classification tasks.

\paragraph{FactorVAE and $\beta$-TCVAE}
\citet{kim2018disentangling} propose a different variant of the VAE objective:
\begin{align}
    L_{\text{FactorVAE}}(p,q) = L_{\text{VAE}}(p,q) - \gamma D_{KL}(q(z)||\prod_{j} q(z_j)). \nonumber
\end{align}
The main idea is to encourage factorization of the aggregate posterior $q(z)=\E_{p^{\text{data}}(x)} \left[ q(z|x) \right]$ so that $z_i$ correlates with $z_j$ if and only if $i=j$.
The authors propose a simple trick to generate samples from the aggregate posterior $q(z)$ and its marginals $\{q(z_j)\}$ using shuffled minibatch indices, then approximate the $D_{KL}(q(z)||\prod_{j} q(z_j))$ term using the cross entropy loss of a classifier that distinguishes between the two sets of samples, which yields a mini-max optimization.

\citet{chen2018isolating} show that the $D_{KL}(q(z)||\prod_{j} q(z_j))$ term above---a.k.a. the ``total correlation'' of the latent code---can be naturally derived by decomposing the expected KL divergence from the variational posterior to prior:
\begin{align} \nonumber
    \E_{p^{\text{data}}(x)}[&D_{KL}(q(z|x)||p(z))] = \\
    &\quad\quad D_{KL}(q(z|x)p^{\text{data}}(x)||q(z)p^{\text{data}}(x)) \nonumber \\
        &\quad\quad+ D_{KL}(q(z)||\prod_j q(z_j))
     \nonumber\\
    &\quad\quad + \sum_j D_{KL} \left[ q(z_j) || p(z_j) \right]. \nonumber
\end{align}
They then augment the decomposed ELBO to arrive at the same objective as \citet{kim2018disentangling}, but optimize using a biased estimate of the marginal probabilities $q(z_j)$ rather than with the adversarial bound on the KL between aggregate posterior and its marginals.

\section{Related Work}
Most work in fair machine learning deals with fairness with respect to single (binary) sensitive attributes. 
Multi-attribute fair classification was recently the focus of \citet{kearns2017preventing}---with empirical follow-up \citep{kearns2018empirical}---and \citet{johnson2018multicalibration}.
Both papers define the notion of an identifiable class of subgroups, and then obtain fair classification algorithms that are provably as efficient as the underlying learning problem for this class of subgroups.
The main difference is the underlying metric; \citet{kearns2017preventing} use statistical parity whereas \citet{johnson2018multicalibration} focus on calibration. 
Building on the multi-accuracy framework of \citet{johnson2018multicalibration}, \citet{kim2018multiaccuracy} develop a new algorithm to achieve multi-group accuracy via a post-processing boosting procedure. 

The search of independent latent components that explain observed data has long been a focus on the probabilistic modeling community \cite{comon1994independent,hyvarinen2000independent,bach2002kernel}.
In light of the increased prevalence of neural networks models in many data domains, the machine learning community has renewed its interest in learned features that ``disentangle'' semantic factors of data variation.
The introduction of the $\beta$-VAE \citep{higgins2016beta}, as discussed in section \ref{sec:background}, motivated a number of subsequent studies  that examine why adding additional weight on the KL-divergence of the ELBO encourages disentangled representations \citep{alemi2018fixing, burgess2018understanding}.
\citet{chen2018isolating,kim2018disentangling} and \citet{esmaeili2018structured} argue that decomposing the ELBO and penalizing the total correlation increases disentanglement in the latent representations. 
\citet{locatello2018challenging} conduct extensive experiments comparing existing unsupervised disentanglement methods and metrics. 
They conclude pessimistically that learning disentangled representations requires inductive biases and possibly additional supervision, but identify fair machine learning as a potential application where additional supervision is available by way of sensitive attributes.

Our work is the first to consider multi-attribute fair representation learning, which we accomplish by using sensitive attributes as labels to induce a factorized structure in the aggregate latent code.
\citet{bose2018compositional} proposed a compositional fair representation of graph-structured data.
\citet{kingma2014semi} previously incorporated (partially-observed) label information into the VAE framework to perform semi-supervised classification.
Several recent VAE variants have incorporated label information into latent variable learning for image synthesis \cite{klys2018learning} and single-attribute fair representation learning \cite{song2018learning,botros2018hierarchical,moyer2018invariant}.
Designing invariant representations with non-variational objectives has also been explored, including in reversible models \cite{ardizzone2018analyzing,jacobsen2018excessive}.

\section{Flexibly Fair VAE} \label{sec:method}

We want to learn fair representations that---beyond being useful for predicting many test-time task labels $y$---can be adapted \textit{simply} and \textit{compositionally} for a variety of sensitive attributes settings $a$ after training.
We call this property \textit{flexible fairness}.
Our approach to this problem involves inducing structure in the latent code that allows for easy manipulation.
Specifically, we isolate information about each sensitive attribute to a specific subspace, while ensuring that the latent space factorizes these subspaces independently. 

\paragraph{Notation}
We employ the following notation:
\begin{itemize}
    \item $x \in \mathcal{X}$: a vector of non-sensitive attributes, for example, the pixel values in an image or row of features in a tabular dataset;
    \item $a \in \{0,1\}^{N_a}$: a vector of binary sensitive attributes;
    \item $z \in \R^{N_z}$: non-sensitive subspace of the latent code;
    \item $b \in \R^{N_b}$: sensitive subspace of the latent code\footnote{
    In our experiments we used $N_b = N_a$ (same number of sensitive attributes as sensitive latent dimensions) to model binary sensitive attributes. 
    But categorical or continuous sensitive attributes can also be accommodated. 
    }.
\end{itemize}
For example, we can express the VAE objective in this notation as
\begin{align}
    L_{\text{VAE}}(p,q) &= \E_{q(z,b|x,a)} \left[ \log p(x,a|z,b) \right] \nonumber \\
    &\quad - D_{KL} \left[ q(z,b|x,a) || p(z,b)  \right]. \nonumber
\end{align}

In learning a flexibly fair representations $[z,b]=f([x,a])$, we aim to satisfy two general properties: \emph{disentanglement} and \emph{predictiveness}.
We say that $[z,b]$ is \emph{disentangled} if its aggregate posterior factorizes as $q(z,b)=q(z)\prod_j q(b_j)$ and is \emph{predictive} if each $b_i$ has high mutual information with the corresponding $a_i$.
Note that under the disentanglement criteria the dimensions of $z$ are free to co-vary together, but must be independent from all sensitive subspaces ${b_j}$.
We have also specified factorization of the latent space in terms of the aggregate posterior $q(z,b)=\E_{p^{\text{data}}(x)} [q(z,b|x)]$, to match the global independence criteria of group fairness.

\paragraph{Desiderata}
We can formally express our desiderata as follows:
\begin{itemize}
    \item $z \perp b_j \medspace \forall \medspace j$ (disentanglement of the non-sensitive and sensitive latent dimensions);
    \item $b_i \perp b_j \medspace \forall \medspace i\neq j$ (disentanglement of the various different sensitive dimensions);
    \item $\text{MI}(a_j,  b_j)$ is large $\forall \medspace j$ (predictiveness of each sensitive dimension);
\end{itemize}
where $\text{MI}(u,v)=\E_{p(u,v)} \log \frac{p(u,v)}{p(u)p(v)}$ represents the mutual information between random vectors $u$ and $v$.
We note that these desiderata differ in two ways from the standard disentanglement criteria.
The predictiveness requirements are stronger: they allow for the injection of external information into the latent representation, requiring the model to structure its latent code to align with that external information.
However, the disentanglement requirement is less restrictive since it allows for correlations between the dimensions of $z$.
Since those are the non-sensitive dimensions, we are not interested in manipulating those at test time, and so we have no need for constraining them.

If we satisfy these criteria, then it is possible to achieve demographic parity with respect to some $a_i$ by simply removing the dimension $b_i$ from the learned representation i.e. use instead $[z,b]\backslash b_i$.
We can alternatively replace $b_i$ with independent noise.
This adaptation procedure is simple and compositional: if we wish to achieve fairness with respect to a conjunction of binary attributes\footnote{
$\wedge$ and $\vee$ represent logical \emph{and} and \emph{or} operations, respectively.
} $a_i \wedge a_j \wedge a_k$, we can simply use the representation $[z,b]\backslash \{b_i, b_j, b_k\}$.

By comparison, while FactorVAE may disentangle dimensions of the aggregate posterior---$q(z)=\prod_j q(z_j)$---it does not automatically satisfy flexible fairness, since the representations are not predictive, and cannot necessarily be easily modified along the attributes of interest.

\paragraph{Distributions}
We propose a variation to the VAE which encourages our desiderata, building on methods for disentanglement and encouraging predictiveness. 
Firstly, we assume assume a variational posterior that factorizes across $z$ and $b$:
\begin{align}
    q(z,b|x) &= q(z|x)q(b|x).
\end{align}
The parameters of these distributions are implemented as neural network outputs, with the encoder network yielding a tuple of parameters for each input: $(\mu_q(x), \Sigma_q(x), \theta_q(x))=\text{Encoder}(x)$.
We then specify $q(z|x)=\mathcal{N}(z|\mu_q(x), \Sigma_q(x))$ and $q(b|x)=\delta(\theta_q(x))$ (i.e., $b$ is non-stochastic)\footnote{
We experimented with several distributions for modeling $b|x$ stochastically, but modeling this uncertainty did not help optimization or downstream evaluation in our experiments.
}.

Secondly, we model reconstruction of $x$ and prediction of $a$ separately using a factorized decoder:
\begin{align}
    p(x,a|z,b)=p(x|z,b)p(a|b)
\end{align}
where $p(x|z,b)$ is the decoder distribution suitably chosen for the inputs $x$, and $p(a|b)=\prod_j\text{Bernoulli}(a_j|\sigma(b_j))$ is a factorized binary classifier that uses $b_j$ as the logit for predicting $a_j$ ($\sigma(\cdot)$ represents the sigmoid function).
Note that the $p(a|b)$ factor of the decoder requires no extra parameters.

Finally, we specify a factorized prior $p(z,b)=p(z)p(b)$ with $p(z)$ as a standard Gaussian and $p(b)$ as Uniform.

\paragraph{Learning Objective}
Using the encoder and decoder as defined above, we present our final objective:
\begin{align} \label{eq:ffvae}
    \nonumber L_{\text{FFVAE}}(p,q) &= \E_{q(z,b|x)} [ \log p(x|z,b) + \alpha \log p(a|b)] \nonumber \\
    &\quad\quad - \gamma D_{KL}(q(z,b)||q(z)\prod_j q(b_j))
     \nonumber\\
    &\quad\quad - D_{KL} \left[ q(z,b|x) || p(z,b) \right].
\end{align}
It comprises the following four terms, respectively:
a \textit{reconstruction} term which rewards the model for successfully modeling non-sensitive observations;
a \textit{predictiveness} term which rewards the model for aligning the correct latent components with the sensitive attributes; 
a \textit{disentanglement} term which rewards the model for decorrelating the latent dimensions of $b$ from each other and $z$;
and a \textit{dimension-wise KL} term which rewards the model for matching the prior in the latent variables.
We call our model \mbox{FFVAE} for Flexibly Fair VAE (see Figure \ref{modelpic} for a schematic representation).

The hyperparameters $\alpha$ and $\gamma$ control aspects relevant to flexible fairness of the representation. $\alpha$ controls the alignment of each $a_j$ to its corresponding $b_j$ (predictiveness), whereas $\gamma$ controls the aggregate independence in the latent code (disentanglement).

The $\gamma$-weighted total correlation term is realized by training a binary adversary  to approximate the log density ratio $\log \frac{q(z,b)}{q(z)\prod_j q(b_j)}$.
The adversary attempts to classify between ``true'' samples from the aggregate posterior $q(z,b)$ and ``fake'' samples from the product of the marginals $q(z)\prod_j q(b_j)$ (see Appendix \ref{sec:disc_approx} for further details).
If a strong adversary can do no better than random chance, then the desired independence property has been achieved.

We note that our model requires the sensitive attributes $a$ at training time but not at test time.
This is advantageous, since often these attributes can be difficult to collect from users, due to practical and legal restrictions, particularly for sensitive information \citep{elliot2008new,division1983}.

\section{Experiments}
\subsection{Evaluation Criteria}

We evaluate the learned encoders with an ``auditing'' scheme on held-out data. 
The overall procedure is as follows:
\begin{enumerate}
    \item \textbf{Split data} into a \textit{training} set (for learning the encoder) and an \textit{audit} set (for evaluating the encoder).
    \item \textbf{Train} an encoder/representation using the training set.
    \item \textbf{Audit} the learned encoder. Freeze the encoder weights and train an MLP to predict some task label given the (possibly modified) encoder outputs on the audit set.
\end{enumerate}

To evaluate various properties of the encoder we conduct three types of auditing tasks---\emph{fair classification}, \emph{predictiveness}, and \emph{disentanglement}---which vary in task label and representation modification.
The \emph{fair classification audit} \citep{madras2018learning} trains an MLP to predict $y$ (held-out from encoder training) given $[z,b]$ with appropriate sensitive dimensions removed, and evaluates accuracy and $\Delta_{DP}$ on a test set.
We repeat for a variety of demographic subgroups derived from the sensitive attributes.
The \emph{predictiveness audit} trains classifier $C_i$ to predict sensitive attribute $a_i$ from $b_i$ alone.
The \emph{disentanglement audit} trains classifier $C_{\backslash i}$ to predict sensitive attribute $a_i$ from the representation with $b_i$ removed (e.g. $[z, b]\backslash b_i$).
If $C_i$ has low loss, our representation is predictive; if $C_{\backslash i}$ has high loss, it is disentangled.

\subsection{Synthetic Data} \label{sec:synthetic}

\newcommand{\figWidth}{0.24\textwidth}
\begin{figure*}[ht!]
%
\begin{subfigure}[t]{\figWidth}
\includegraphics[width=\textwidth]{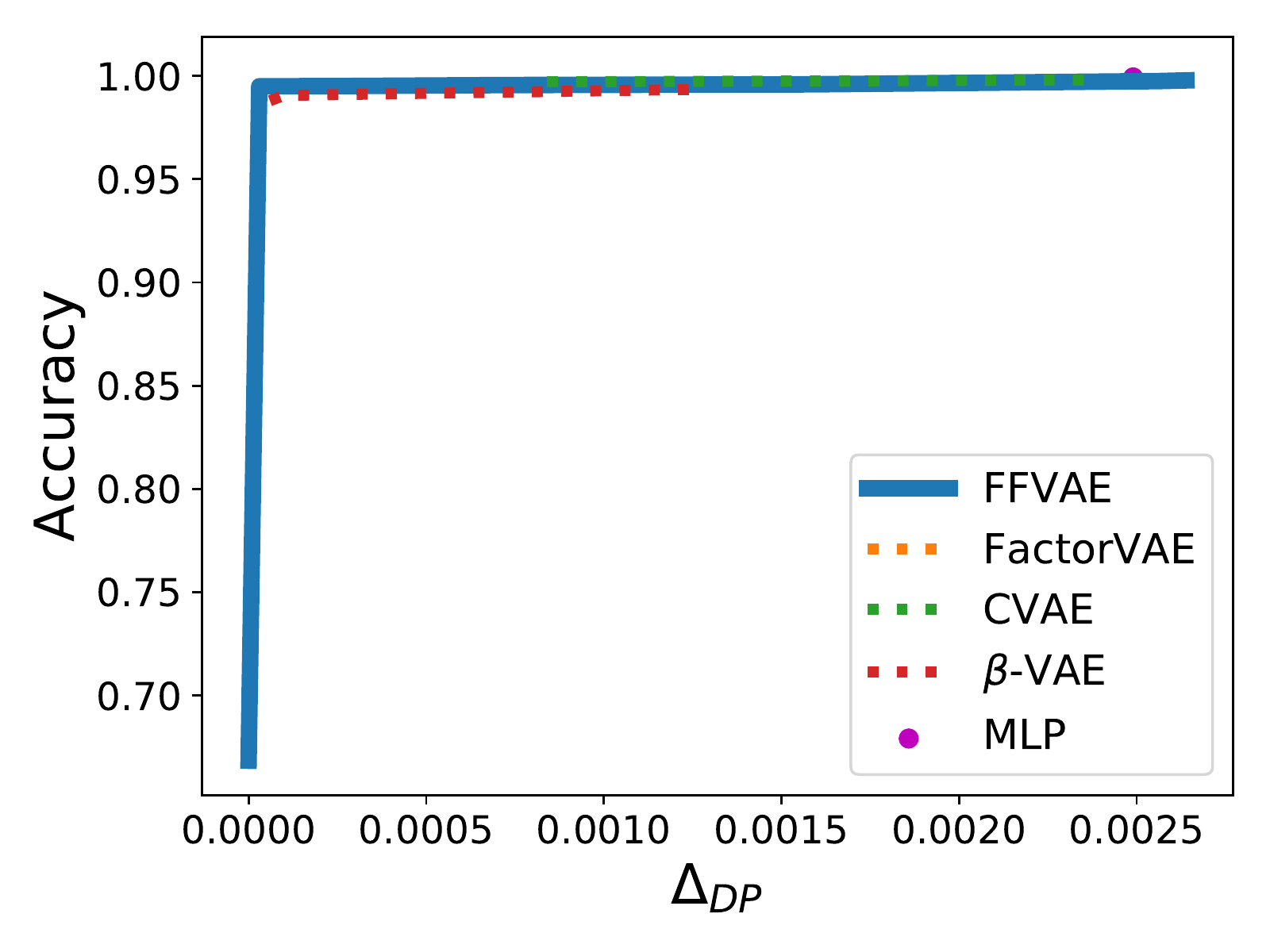}
\caption{$a$ = Scale}
\label{fig:dsprites-pareto-a2-y4}
\end{subfigure}
\hfill
\begin{subfigure}[t]{\figWidth}
\includegraphics[width=\textwidth]{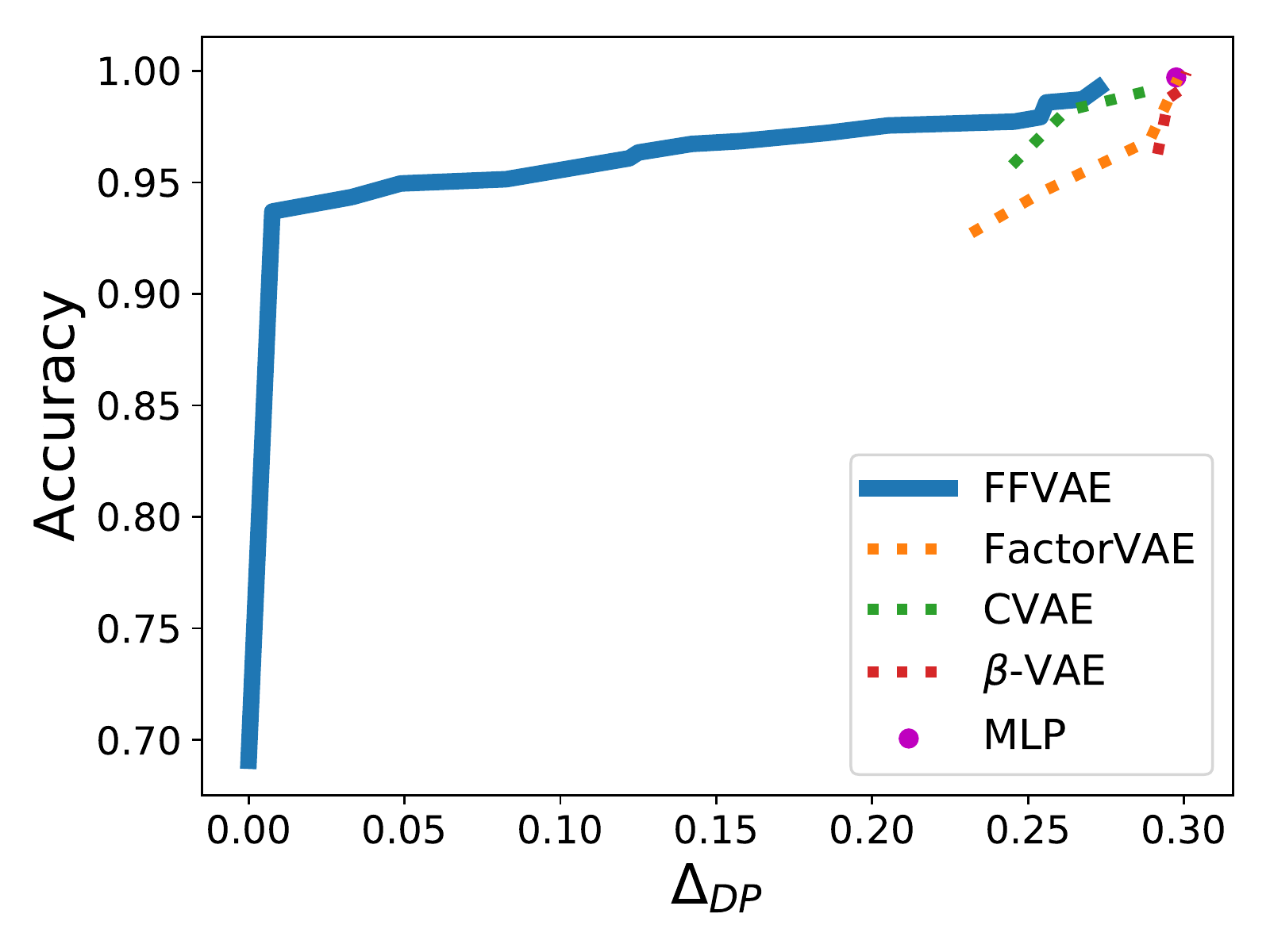}
\caption{$a$ = Shape}
\label{fig:dsprites-pareto-a1-y4}
\end{subfigure}
\hfill
\begin{subfigure}[t]{\figWidth}
\includegraphics[width=\textwidth]{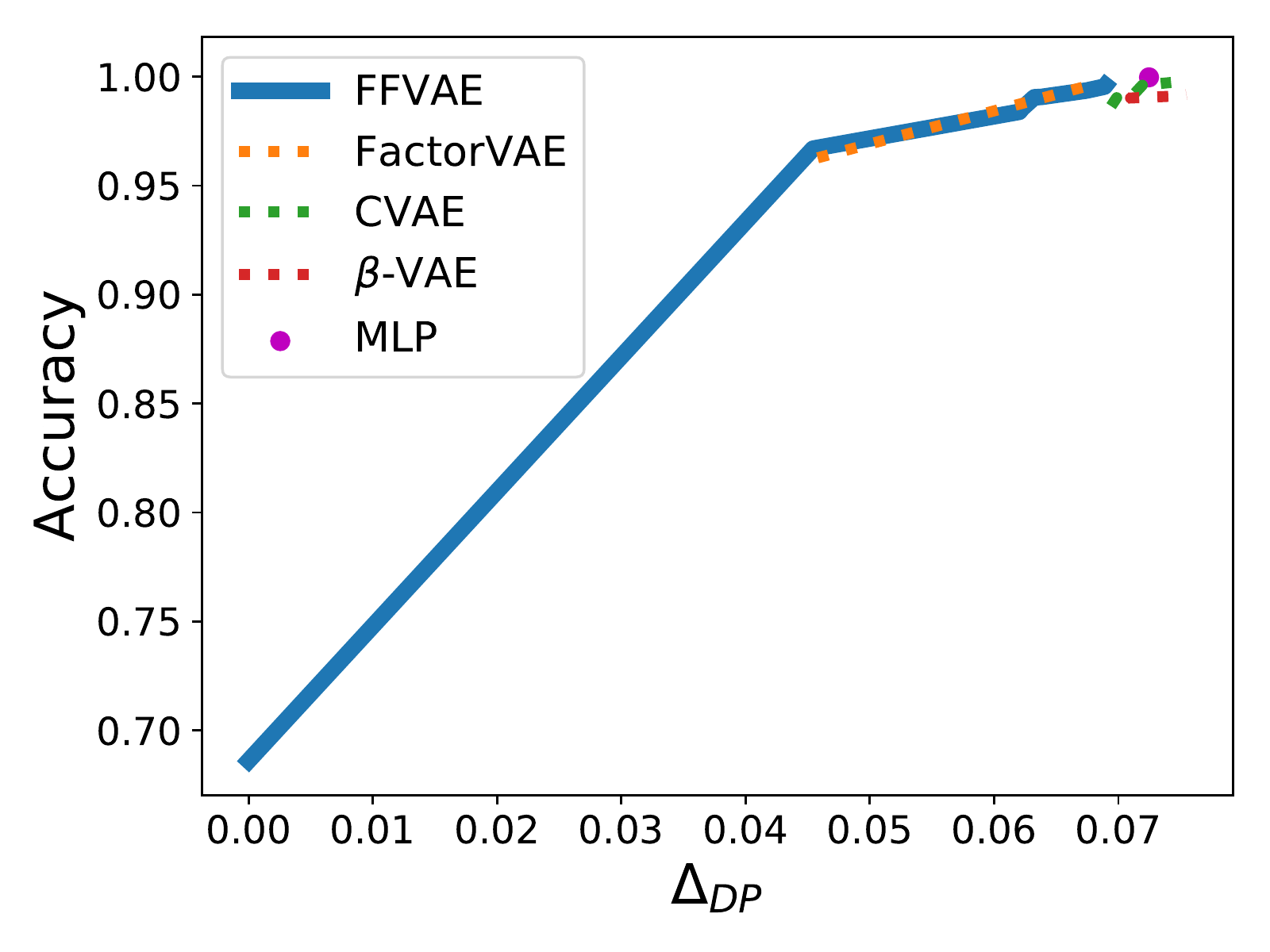}    
\caption{$a$ = Shape $\wedge$ Scale}
\label{fig:dsprites-pareto-a1and2-y4}
\end{subfigure}
\hfill
\begin{subfigure}[t]{\figWidth}
\includegraphics[width=\textwidth]{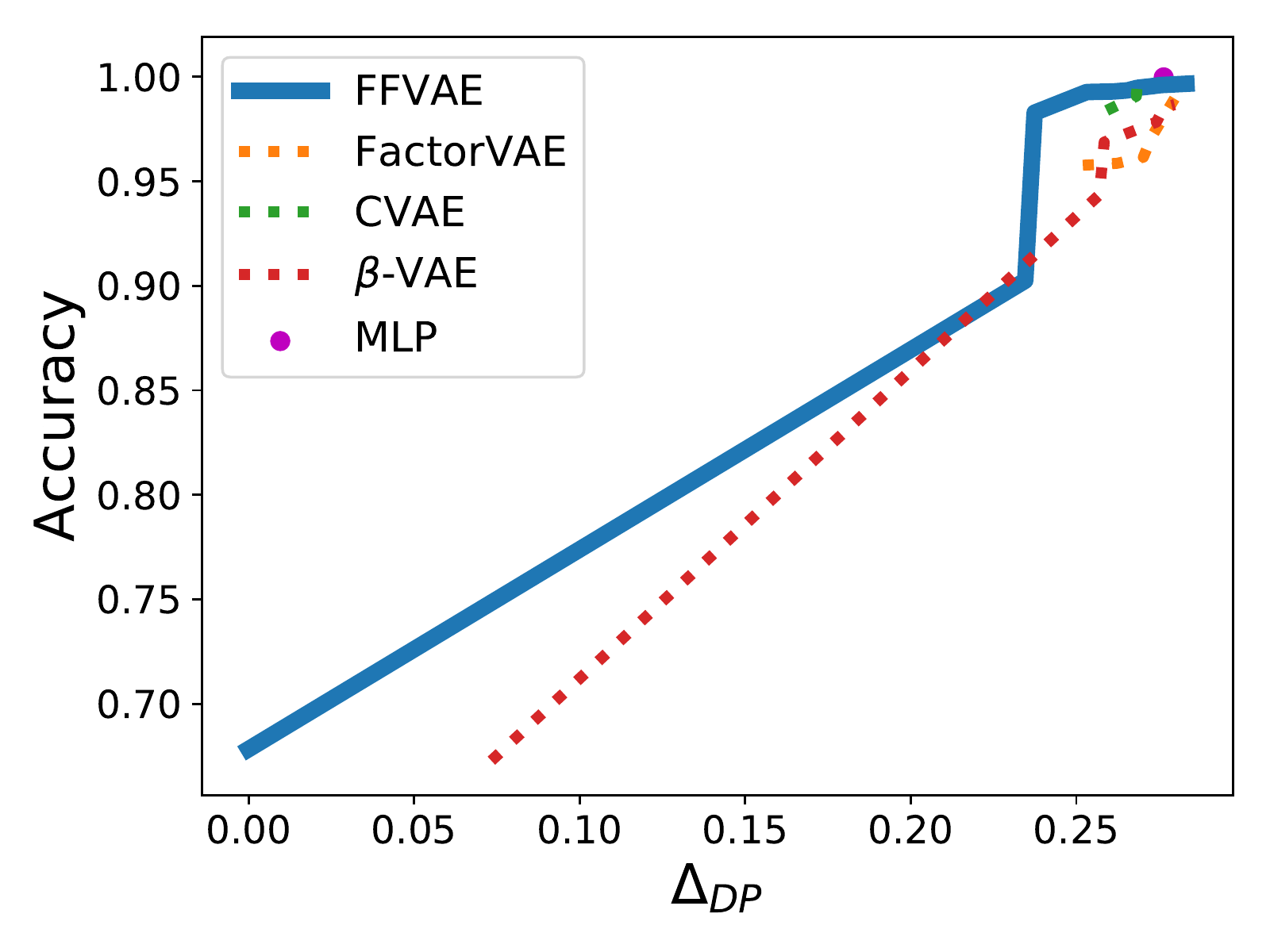}
\caption{$a$ = Shape $\vee$ Scale}
\label{fig:dsprites-pareto-a1or2-y4}
\end{subfigure}

\caption{
    Fairness-accuracy tradeoff curves, DSpritesUnfair dataset.
    We sweep a range of hyperparameters for each model and report Pareto fronts.
    Optimal point is the top left hand corner --- this represents perfect accuracy and fairness.
    MLP is a baseline classifier trained directly on the input data.
    For each model, encoder outputs are modified to remove information about $a$.
    $y$ = XPosition for each plot.
    }
    \label{fig:dsprites-pareto}
\end{figure*}

\paragraph{DSpritesUnfair Dataset}
\begin{figure}[ht!]
\centering
\includegraphics[width=0.4\textwidth]{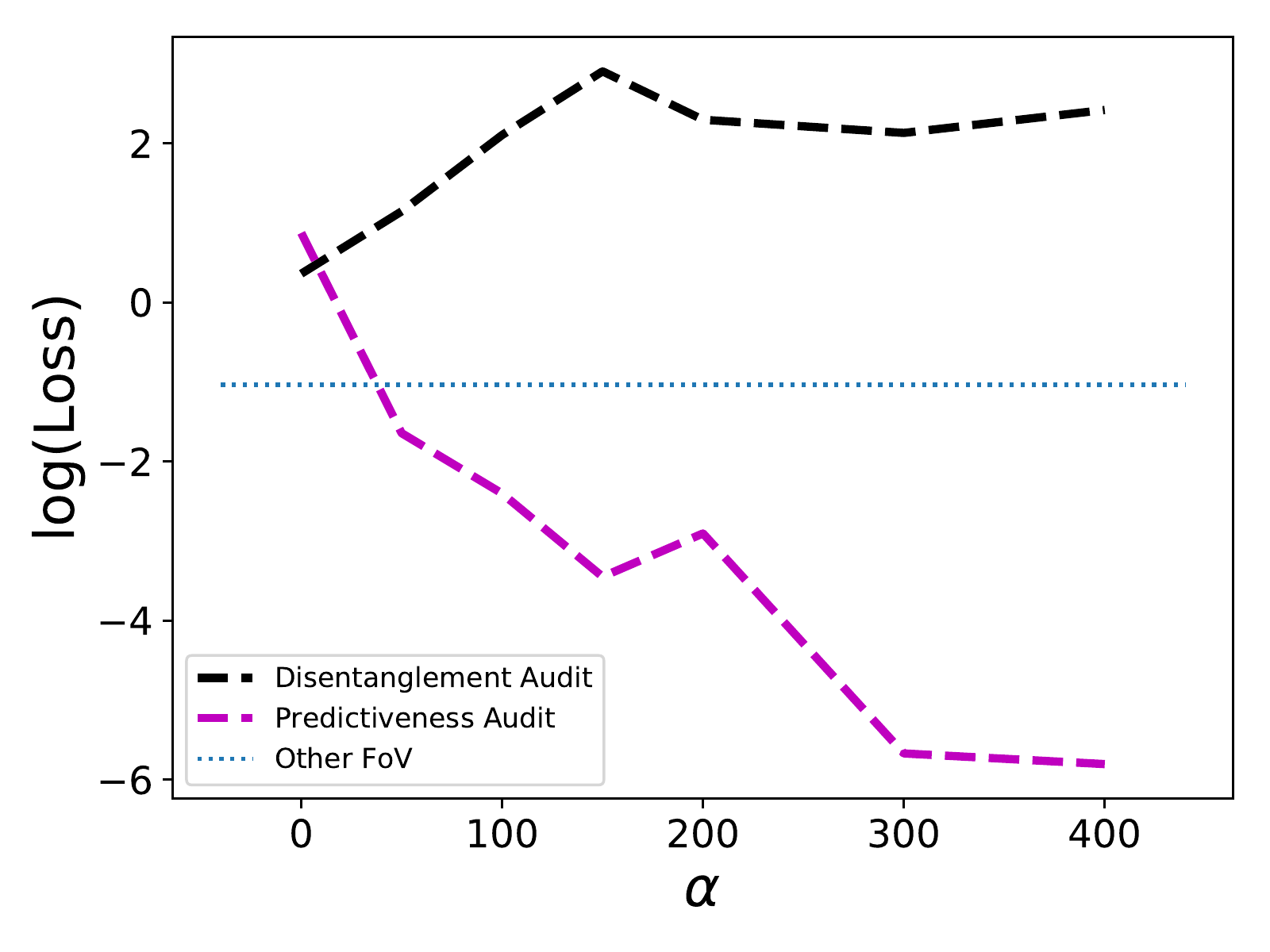}
\vspace{-0.2cm}
\caption{
    Black and pink dashed lines respectively show FFVAE disentanglement audit (the higher the better) and predictiveness audit (the lower the better) as a function of $\alpha$.
    These audits use $A_i$=Shape (see text for details).
    The blue line is a reference value---the log loss of a classifier that predicts $A_i$ from the other 5 DSprites factors of variation (FoV) alone, ignoring the image---representing the amount of information about $A_i$ inherent in the data.
    }
    \label{fig:dsprites-audit}
\end{figure}

The DSprites dataset\footnote{\scriptsize \url{https://github.com/deepmind/dsprites-dataset}} contains $64 \times 64$-pixel images of white shapes against a black background, and was designed to evaluate whether learned representations have disentangled sources of variation.
The original dataset has several categorical factors of variation---Scale, Orientation, XPosition, YPosition---that combine to create $700,000$ unique images.
We binarize the factors of variation to derive sensitive attributes and labels, so that many images now share any given attribute/label combination (See Appendix \ref{sec:dsprites-generation} for details).
In the original DSprites dataset, the factors of variation are sampled uniformly.
However, in fairness problems, we are often concerned with correlations between attributes and the labels we are trying to predict (otherwise, achieving low $\Delta_{DP}$ is aligned with standard classification objectives).
Hence, we sampled an ``unfair'' version of this data (DSpritesUnfair) with correlated factors of variation; in particular Shape and XPosition correlate positively.
Then a non-trivial fair classification task would be, for instance, learning to predict shape without discriminating against inputs from the left side of the image.

\paragraph{Baselines}
To test the utility of our predictiveness prior, we compare our model to $\beta$-VAE (VAE with a coefficient $\beta \geq 1$ on the KL term) and FactorVAE, which have disentanglement priors but no predictiveness prior.
We can also think of these as FFVAE with $\alpha = 0$.
To test the utility of our disentanglement prior, we also compare against a version of our model with $\gamma = 0$, denoted CVAE.
This is similar to the class-conditional VAE \citep{kingma2014semi}, with sensitive attributes as labels --- this model encourages predictiveness but no disentanglement.

\paragraph{Fair Classification}
We perform the fair classification audit using several group/subgroup definitions for models trained on DSpritesUnfair (see Appendix \ref{sec:dsprites-training-details} for training details), and report fairness-accuracy tradeoff curves in Fig. \ref{fig:dsprites-pareto}.
In these experiments, we used Shape and Scale as our sensitive attributes during encoder training.
We perform the fair classification audit by training an MLP to predict $y=$``XPosition''
---which was not used in the representation learning phase---given the modified encoder outputs, and repeat for several sensitive groups and subgroups.
We modify the encoder outputs as follows:
When our sensitive attribute is $a_i$ we remove the associated dimension $b_i$ from $[z,b]$; when the attribute is a conjunction of $a_i$ and $a_j$, we remove both $b_i$ and $b_j$.
For the baselines, we simply remove the latent dimension which is most correlated with $a_i$, or the two most correlated dimensions with the conjunction.
We sweep a range of hyperparameters to produce the fairness-accuracy tradeoff curve for each model.
In Fig. \ref{fig:dsprites-pareto}, we show the ``Pareto front'' of these models: points in ($\Delta_{DP}$, accuracy)-space for which no other point is better along both dimensions.
The optimal result is the top left hand corner (perfect accuracy and $\Delta_{DP} = 0$).

Since we have a 2-D sensitive input space, we show results for four different sensitive attributes derived from these dimensions: $\{a=\text{``Shape''}, a=\text{``Scale''}, a=\text{``Shape''}\vee\text{``Scale''}, a=\text{``Shape''}\wedge\text{``Scale''}\}$.
Recall that Shape and XPosition correlate in the DSpritesUnfair dataset.
Therefore, for sensitive attributes that involve Shape, we expect to see an improvement in $\Delta_{DP}$.
For sensitive attributes that do not involve Shape, we expect that our method does not hurt performance at all --- since the attributes are uncorrelated in the data, the optimal predictive solution also has $\Delta_{DP} = 0$.

When group membership $a$ is uncorrelated with label $y$ (Fig. \ref{fig:dsprites-pareto-a2-y4}), all models achieve high accuracy and low $\Delta_{DP}$ ($a$ and $y$ successfully disentangled).
When $a$ correlates with $y$ by design (Fig. \ref{fig:dsprites-pareto-a1-y4}), we see the clearest improvement of the FFVAE over the baselines, with an almost complete reduction in $\Delta_{DP}$ and very little accuracy loss.
The baseline models are all unable to improve $\Delta_{DP}$ by more than about 0.05, indicating that they have not effectively disentangled the sensitive information from the label.
In Figs. \ref{fig:dsprites-pareto-a1and2-y4} and \ref{fig:dsprites-pareto-a1or2-y4}, we examine conjunctions of sensitive attributes, assessing FFVAE's ability to flexibly provide multi-attribute fair representations.
Here FFVAE exceeds or matches the baselines accuracy-at-a-given-$\Delta_{DP}$ almost everywhere;
by disentangling information from multiple sensitive attributes, FFVAE enables flexibly fair downstream classification.

\paragraph{Disentanglement and Predictiveness}
Fig. \ref{fig:dsprites-audit} shows the FFVAE disentanglement and predictiveness audits (see above for description of this procedure).
This result aggregates audits across all FFVAE models trained in the setting from Figure \ref{fig:dsprites-pareto-a1-y4}.
The classifier loss is cross-entropy, which is a lower bound on the mutual information between the input and target of the classifier.
We observe that increasing $\alpha$ helps both predictiveness and disentanglement in this scenario.
In the disentanglement audit, larger $\alpha$ makes predicting the sensitive attribute from the modified representation (with $b_i$ removed) more difficult.
The horizontal dotted line shows the log loss of a classifier that predicts $a_i$ from the other DSprites factors of variation (including labels not available to FFVAE); this baseline reflects the correlation inherent in the data.
We see that when $\alpha = 0$ (i.e. FactorVAE), it is slightly more difficult than this baseline to predict the sensitive attribute.
This is due to the disentanglement prior.
However, increasing $\alpha > 0$ increases disentanglement benefits in FFVAE beyond what is present in FactorVAE.
This shows that encouraging predictive structure can help disentanglement through isolating each attribute's information in particular latent dimensions.
Additionally, increasing $\alpha$ improves predictiveness, as expected from the objective formulation.
We further evaluate the disentanglement properties of our model in Appendix \ref{sec:mig} using the Mutual Information Gap metric \cite{chen2018isolating}.

\subsection{Communities \& Crime} \label{sec:communities}
\paragraph{Dataset}
Communities \& Crime\footnote{\scriptsize \url{http://archive.ics.uci.edu/ml/datasets/communities+and+crime}} is a tabular UCI dataset containing neighborhood-level population statistics. 
120 such statistics are recorded for each of the $1,994$ neighborhoods. 
Several attributes encode demographic information that may be protected. We chose three as sensitive: racePctBlack (\% neighborhood population which is Black), blackPerCap (avg per capita income of Black residents), and pctNotSpeakEnglWell (\% neighborhood population that does not speak English well). 
We follow the same train/eval procedure as with DSpritesUnfair - we train FFVAE with the sensitive attributes and evaluate using naive MLPs to predict a held-out label (violent crimes per capita) on held-out data.

\newcommand{\thirdFigWidth}{0.15\textwidth}

\begin{figure}[ht!]
%
\begin{subfigure}[t]{\thirdFigWidth}
\includegraphics[width=\textwidth]{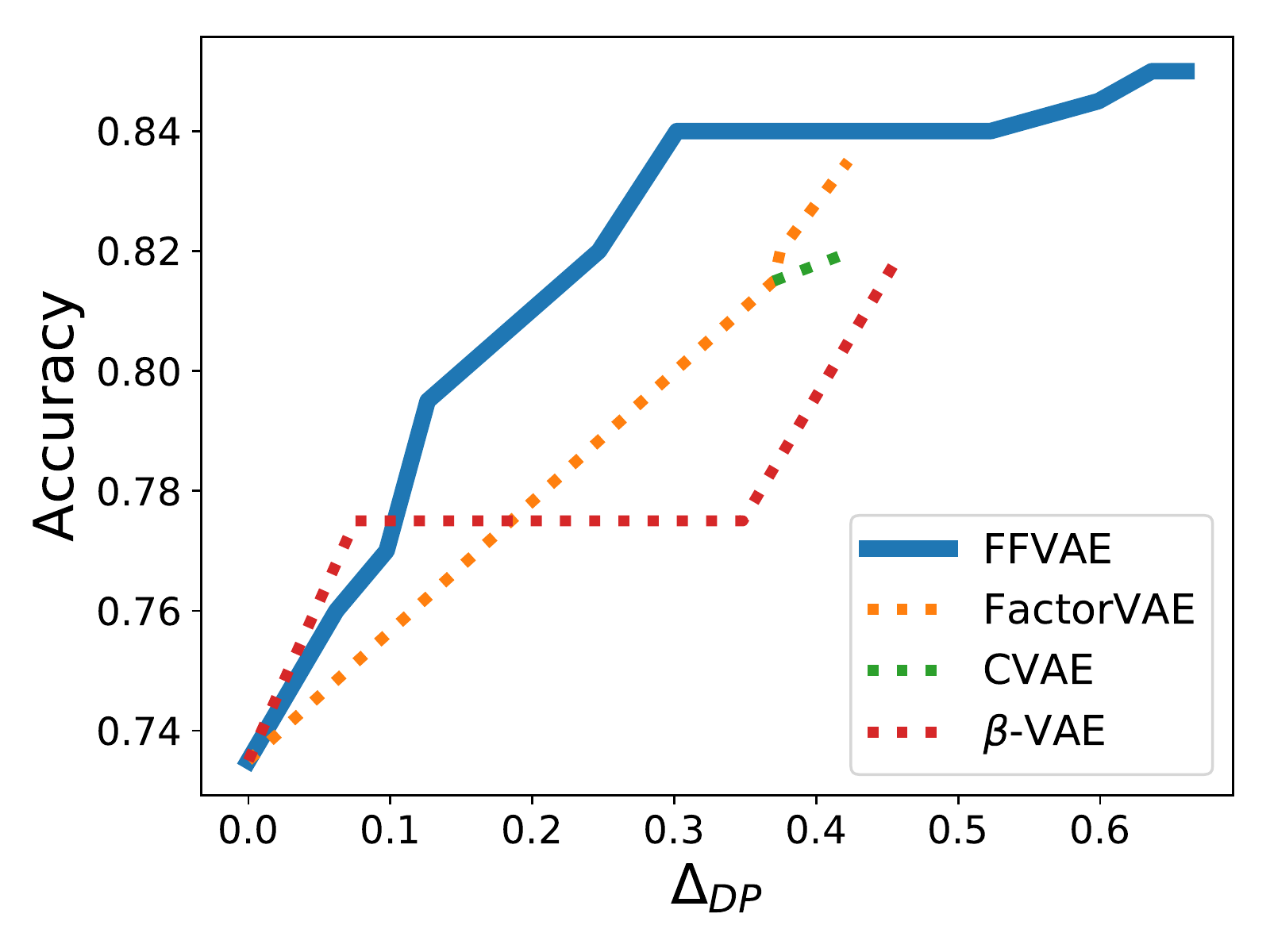}
\caption{$a$ = R}
\label{fig:cc_attr_fn_0}
\end{subfigure}
\hfill
\begin{subfigure}[t]{\thirdFigWidth}
\includegraphics[width=\textwidth]{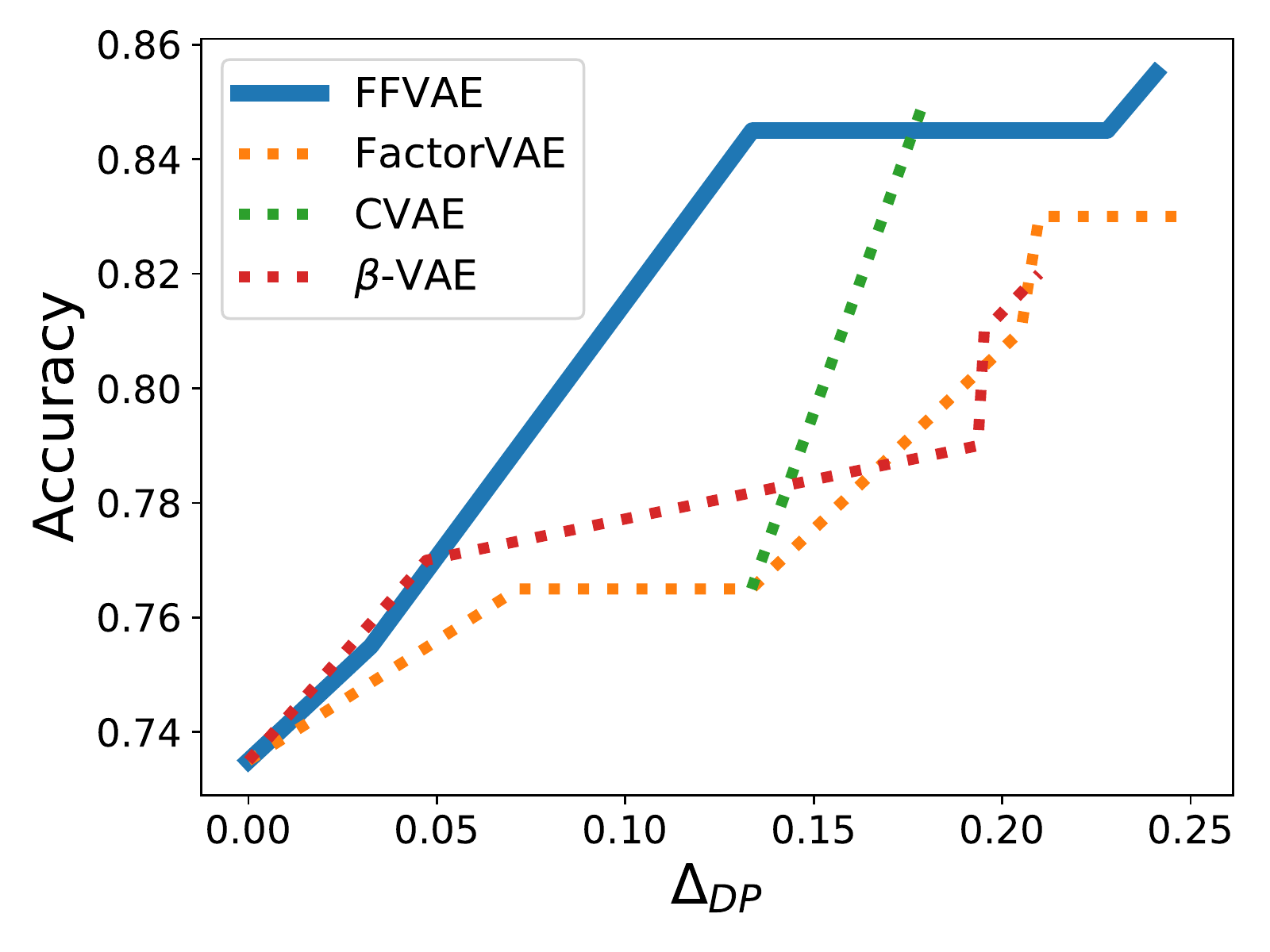}
\caption{$a$ = B}
\label{fig:cc_attr_fn_5}
\end{subfigure}
\hfill
\begin{subfigure}[t]{\thirdFigWidth}
\includegraphics[width=\textwidth]{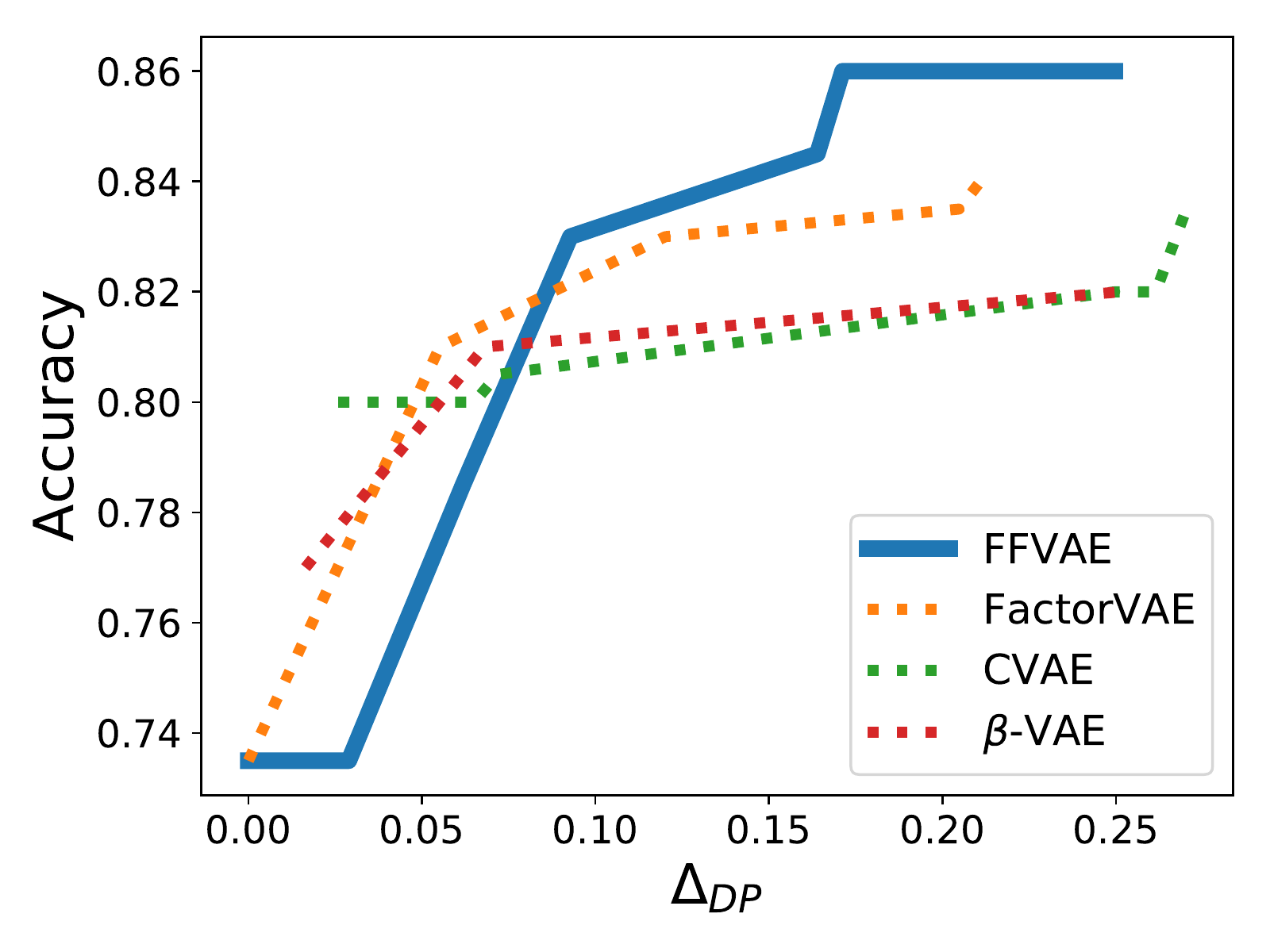}
\caption{$a$ = P}
\label{fig:cc_attr_fn_11}
\end{subfigure}
\hfill
\begin{subfigure}[t]{\thirdFigWidth}
\includegraphics[width=\textwidth]{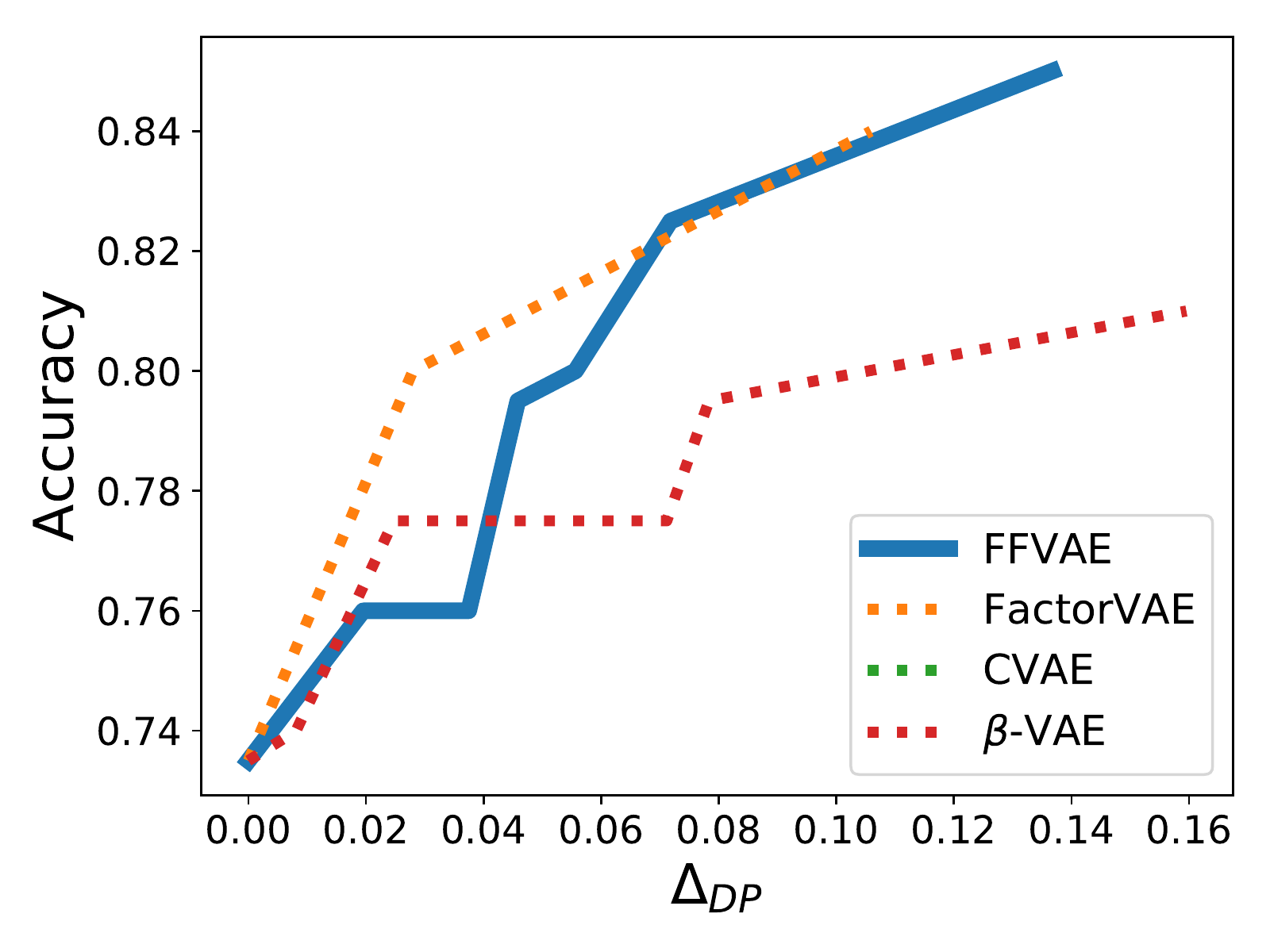}
\caption{$a$ = R $\vee$ B}
\label{fig:cc_attr_fn_0_OR_cc_attr_fn_5}
\end{subfigure}
\hfill
\begin{subfigure}[t]{\thirdFigWidth}
\includegraphics[width=\textwidth]{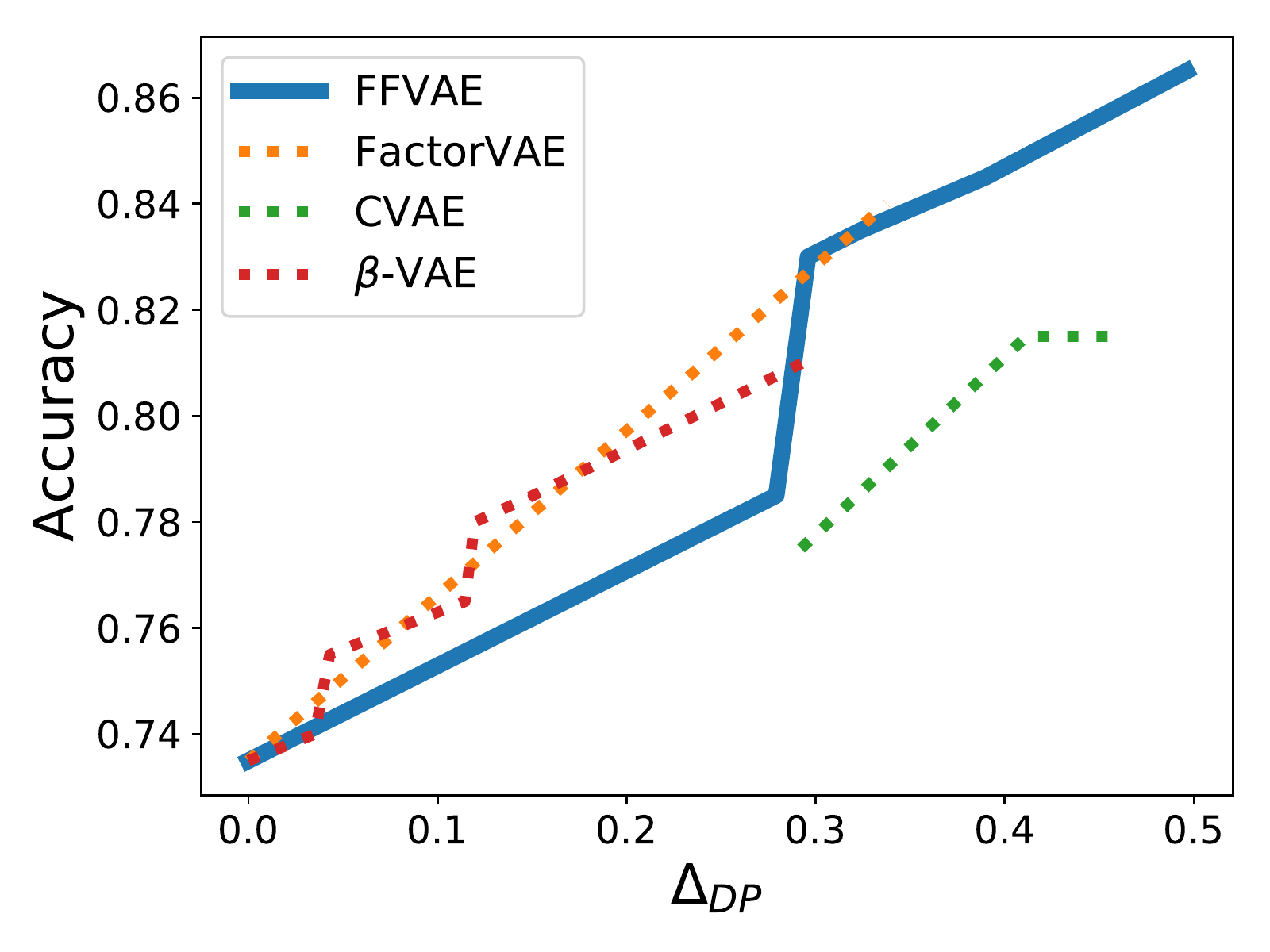}
\caption{$a$ = R $\vee$ P}
\label{fig:cc_attr_fn_0_OR_cc_attr_fn_11}
\end{subfigure}
\hfill
\begin{subfigure}[t]{\thirdFigWidth}
\includegraphics[width=\textwidth]{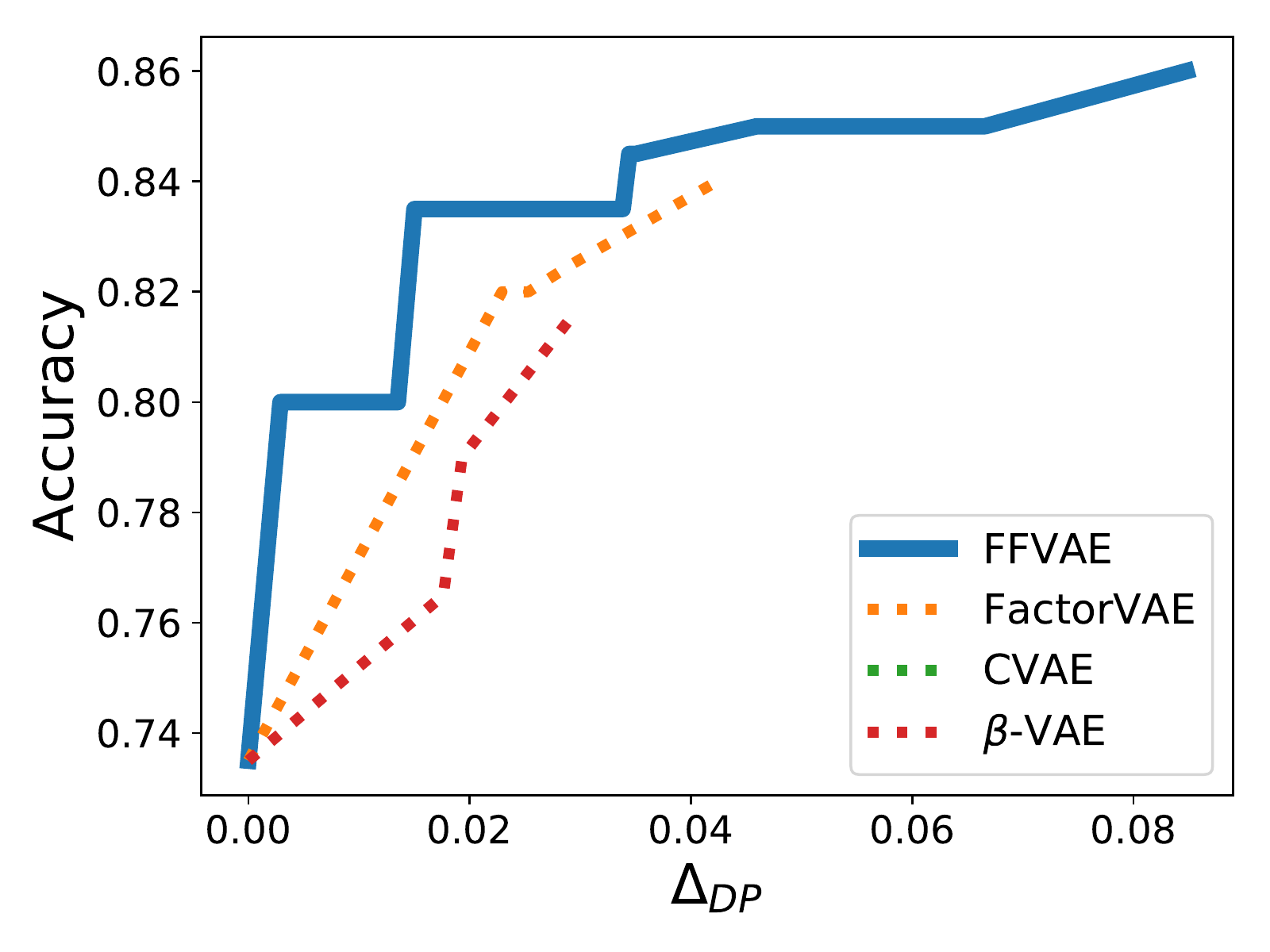}
\caption{$a$ = B $\vee$ P}
\label{fig:cc_attr_fn_5_OR_cc_attr_fn_11}
\end{subfigure}
\hfill
\begin{subfigure}[t]{\thirdFigWidth}
\includegraphics[width=\textwidth]{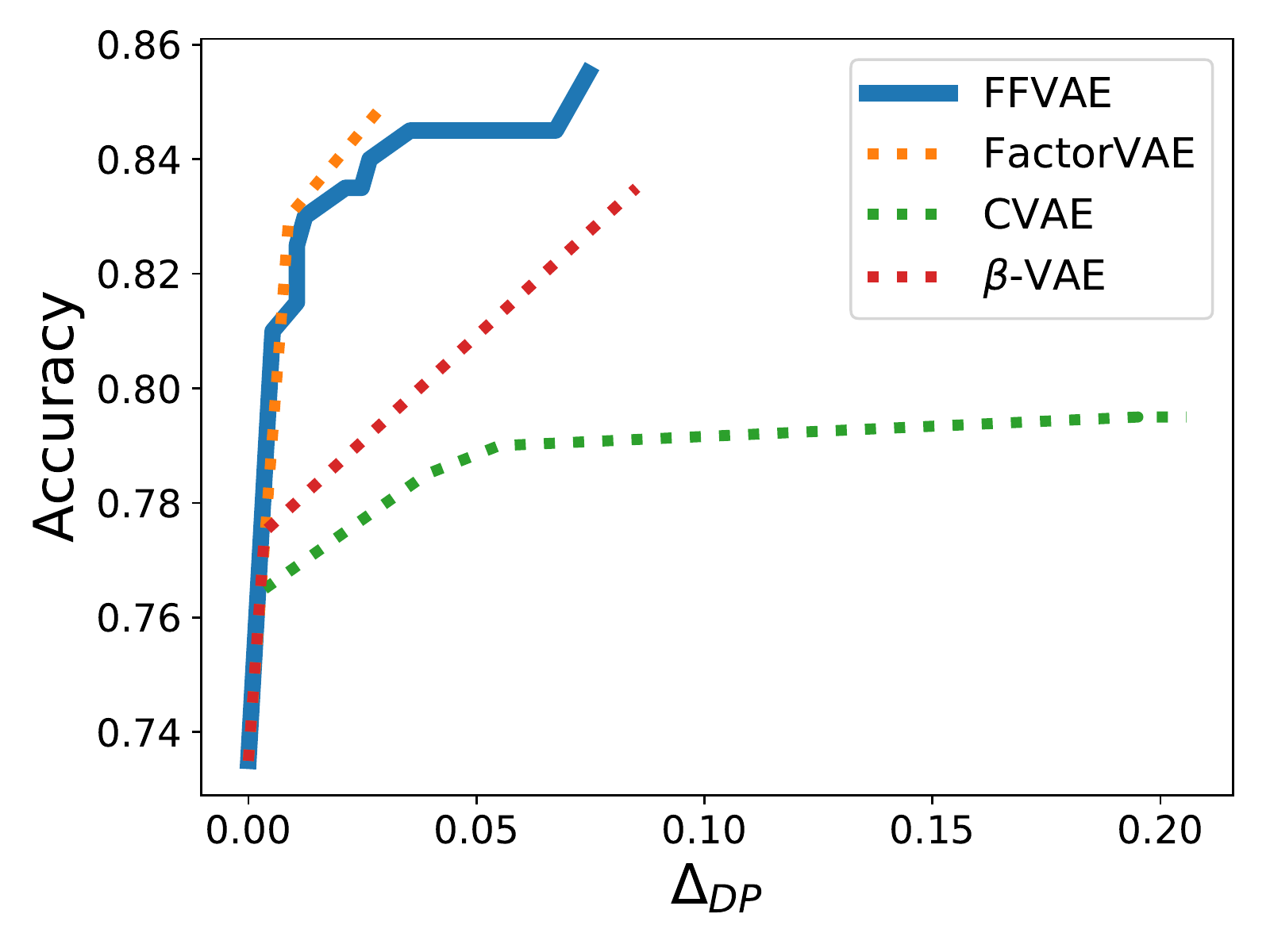}
\caption{$a$ = R $\wedge$ B}
\label{fig:cc_attr_fn_0_AND_cc_attr_fn_5}
\end{subfigure}
\hfill
\begin{subfigure}[t]{\thirdFigWidth}
\includegraphics[width=\textwidth]{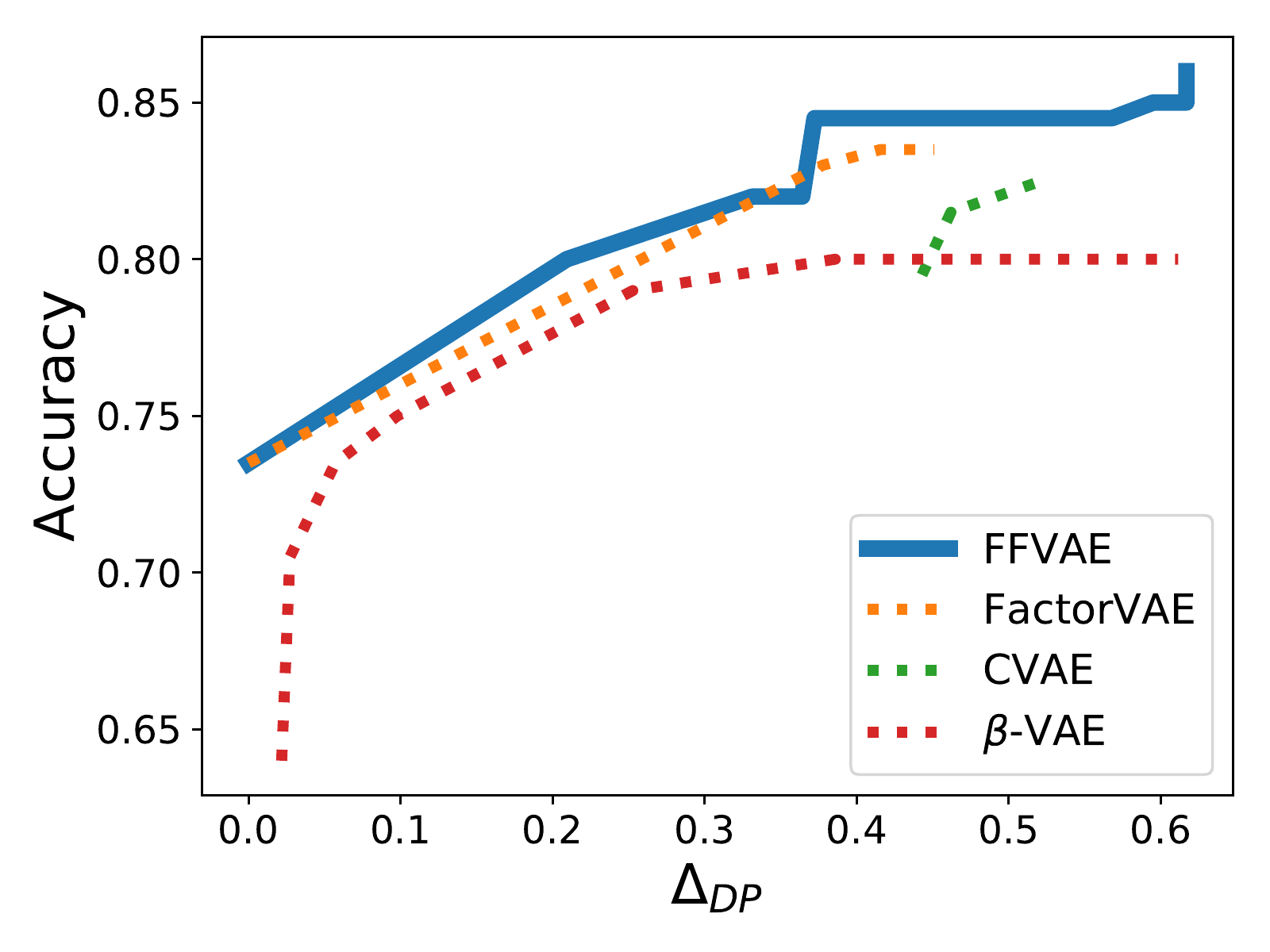}
\caption{$a$ = R $\wedge$ P}
\label{fig:cc_attr_fn_0_AND_cc_attr_fn_11}
\end{subfigure}
\hfill
\begin{subfigure}[t]{\thirdFigWidth}
\includegraphics[width=\textwidth]{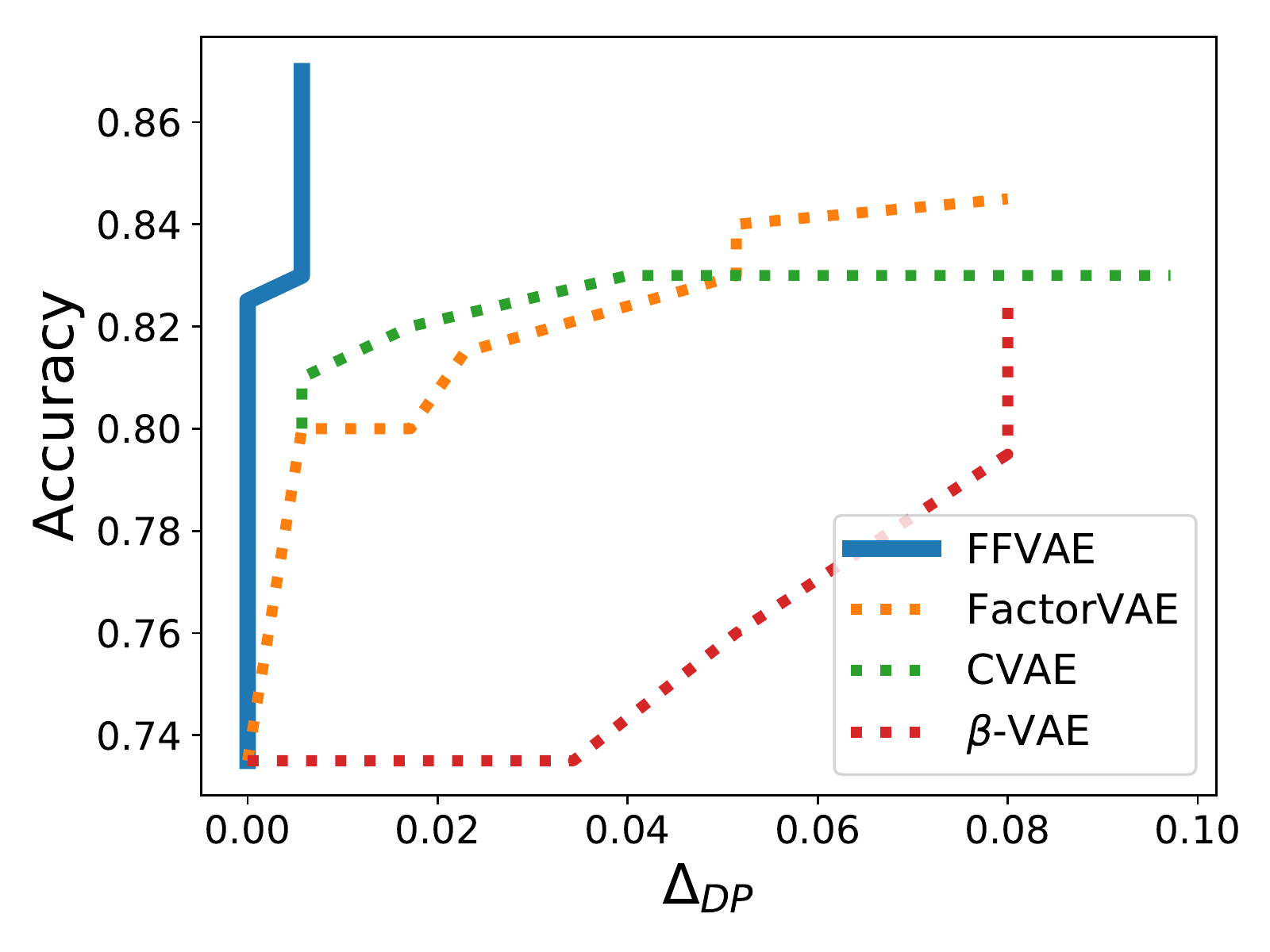}
\caption{$a$ = B $\wedge$ P}
\label{fig:cc_attr_fn_5_AND_cc_attr_fn_11}
\end{subfigure}
\hfill

\caption{
    Communities \& Crime subgroup fairness-accuracy tradeoffs.
    Sensitive attributes: racePctBlack (R), blackPerCapIncome (B), and pctNotSpeakEnglWell (P).
    $y$ = violentCrimesPerCaptia. 
    }
    \label{fig:comcrime-pareto}
\end{figure}

\paragraph{Fair Classification}
This dataset presents a more difficult disentanglement problem than DSpritesUnfair.
The three sensitive attributes we chose in Communities and Crime were somewhat correlated with each other, a natural artefact of using real (rather than simulated) data.
We note that in general, the disentanglement literature does not provide much guidance in terms of disentangling correlated attributes.
Despite this obstacle, FFVAE performed reasonably well in the fair classification audit (Fig. \ref{fig:comcrime-pareto}).
It achieved higher accuracy than the baselines in general, likely due to its ability to incorporate side information from $a$ during training.
Among the baselines, FactorVAE tended perform best, suggesting achieving a factorized aggregate posterior helps with fair classification.
While our method does not outperform the baselines on each conjunction, its relatively strong performance on a difficult, tabular dataset shows the promise of using disentanglement priors in designing robust subgroup-fair machine learning models.

\subsection{Celebrity Faces} \label{sec:celeba}

\begin{figure}[ht!]
%
\begin{subfigure}[t]{\thirdFigWidth}
\includegraphics[width=\textwidth]{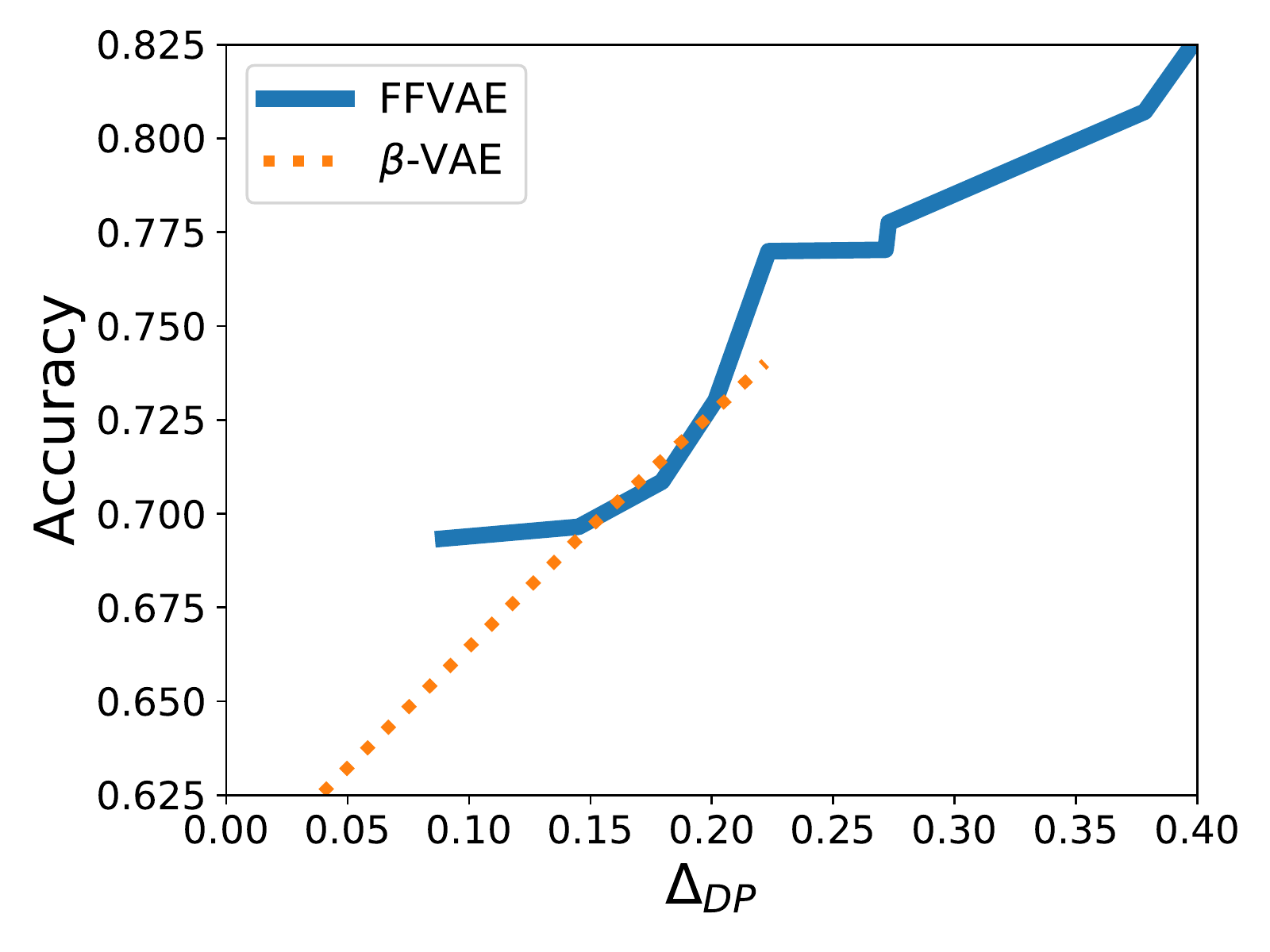}
\caption{$a$ = C}
\label{fig:predict_C}
\end{subfigure}
\hfill
\begin{subfigure}[t]{\thirdFigWidth}
\includegraphics[width=\textwidth]{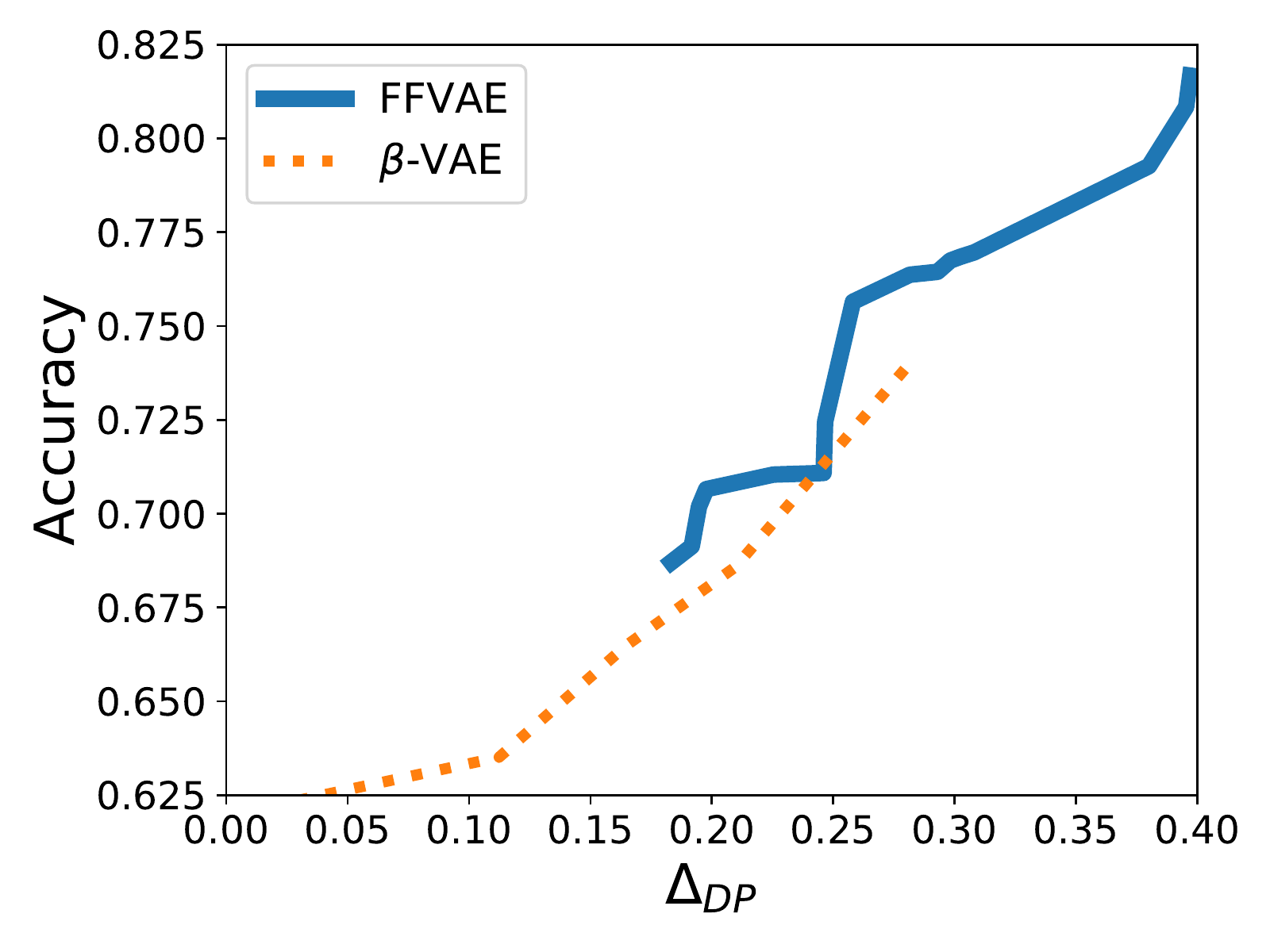}
\caption{$a$ = E}
\label{fig:predict_E}
\end{subfigure}
\hfill
\begin{subfigure}[t]{\thirdFigWidth}
\includegraphics[width=\textwidth]{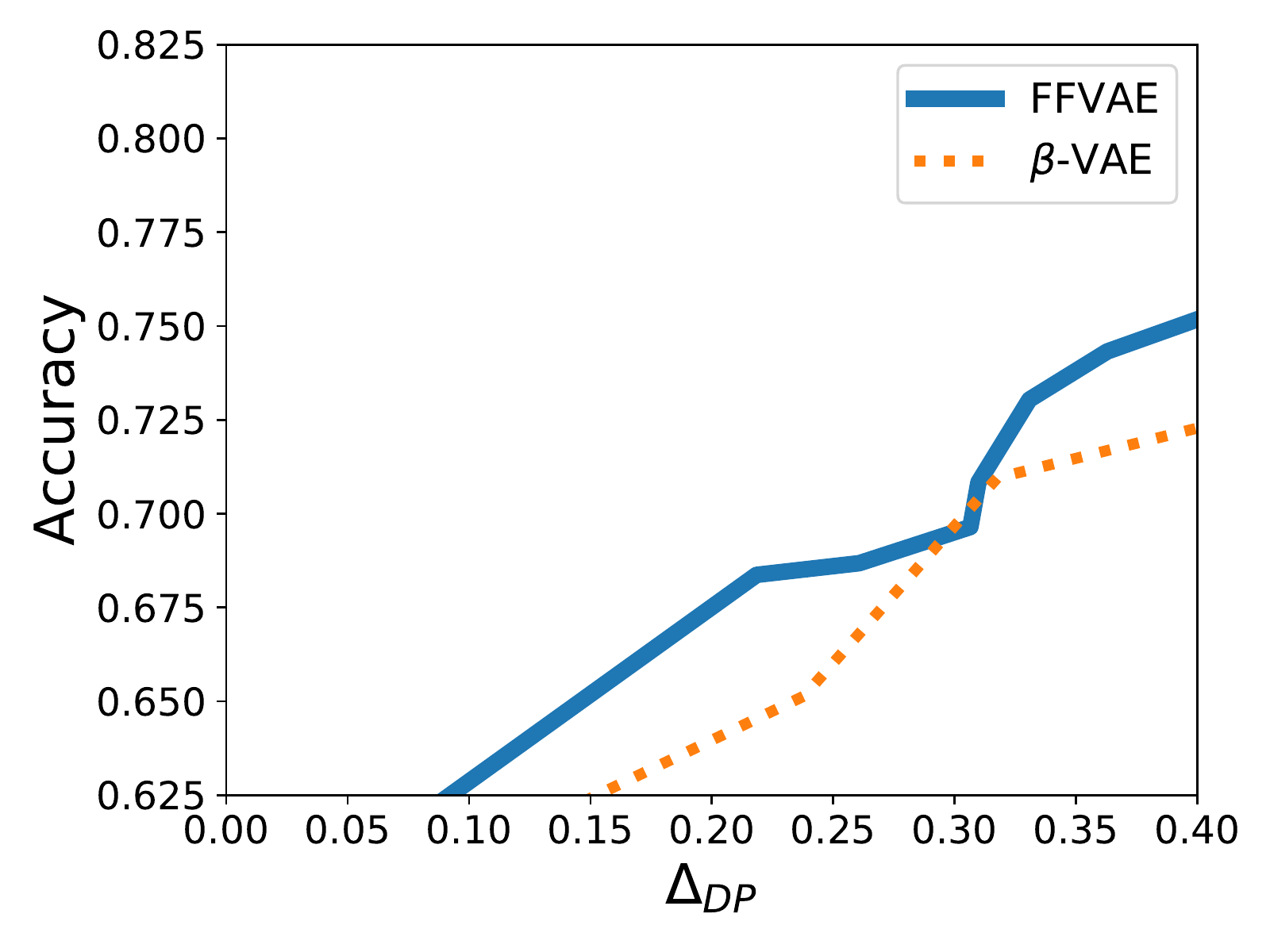}
\caption{$a$ = M}
\label{fig:predict_M}
\end{subfigure}
\hfill
\begin{subfigure}[t]{\thirdFigWidth}
\includegraphics[width=\textwidth]{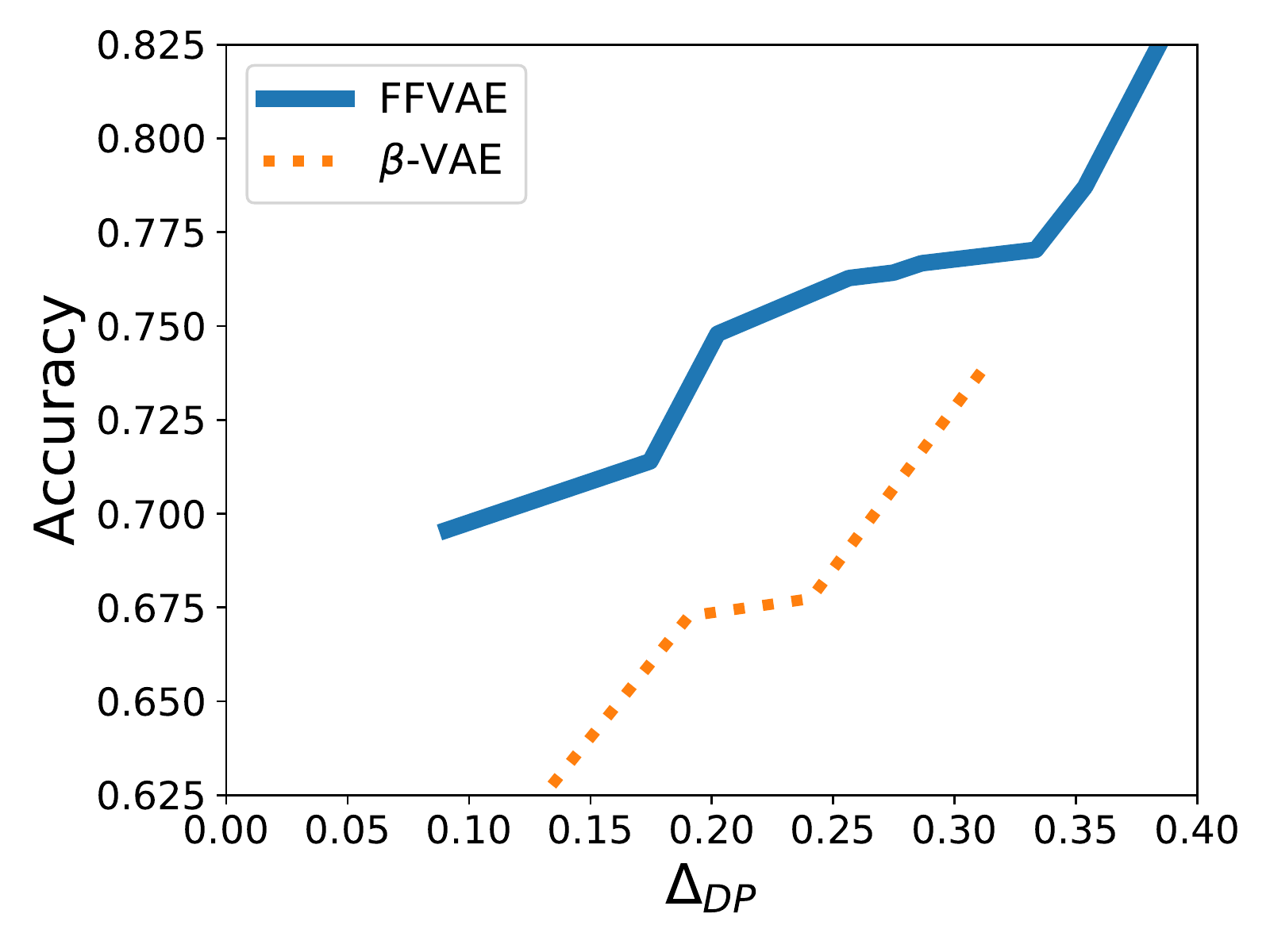}
\caption{$a$ = C $\wedge$ E}
\label{fig:predict_C-AND-E}
\end{subfigure}
\hfill
\begin{subfigure}[t]{\thirdFigWidth}
\includegraphics[width=\textwidth]{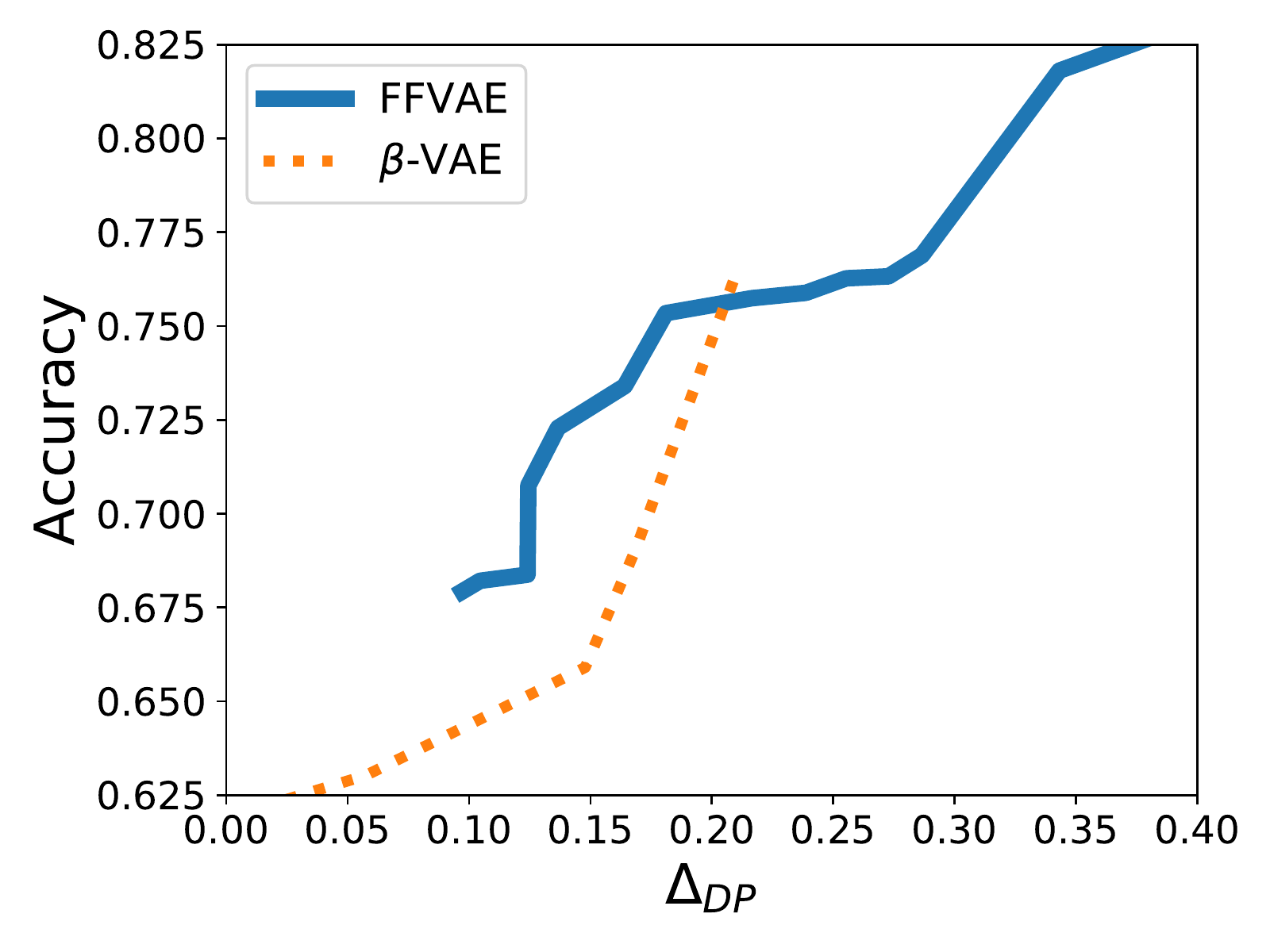}
\caption{$a$ = C $\wedge \neg$ E}
\label{fig:predict_C-AND-NOT-E}
\end{subfigure}
\hfill
\begin{subfigure}[t]{\thirdFigWidth}
\includegraphics[width=\textwidth]{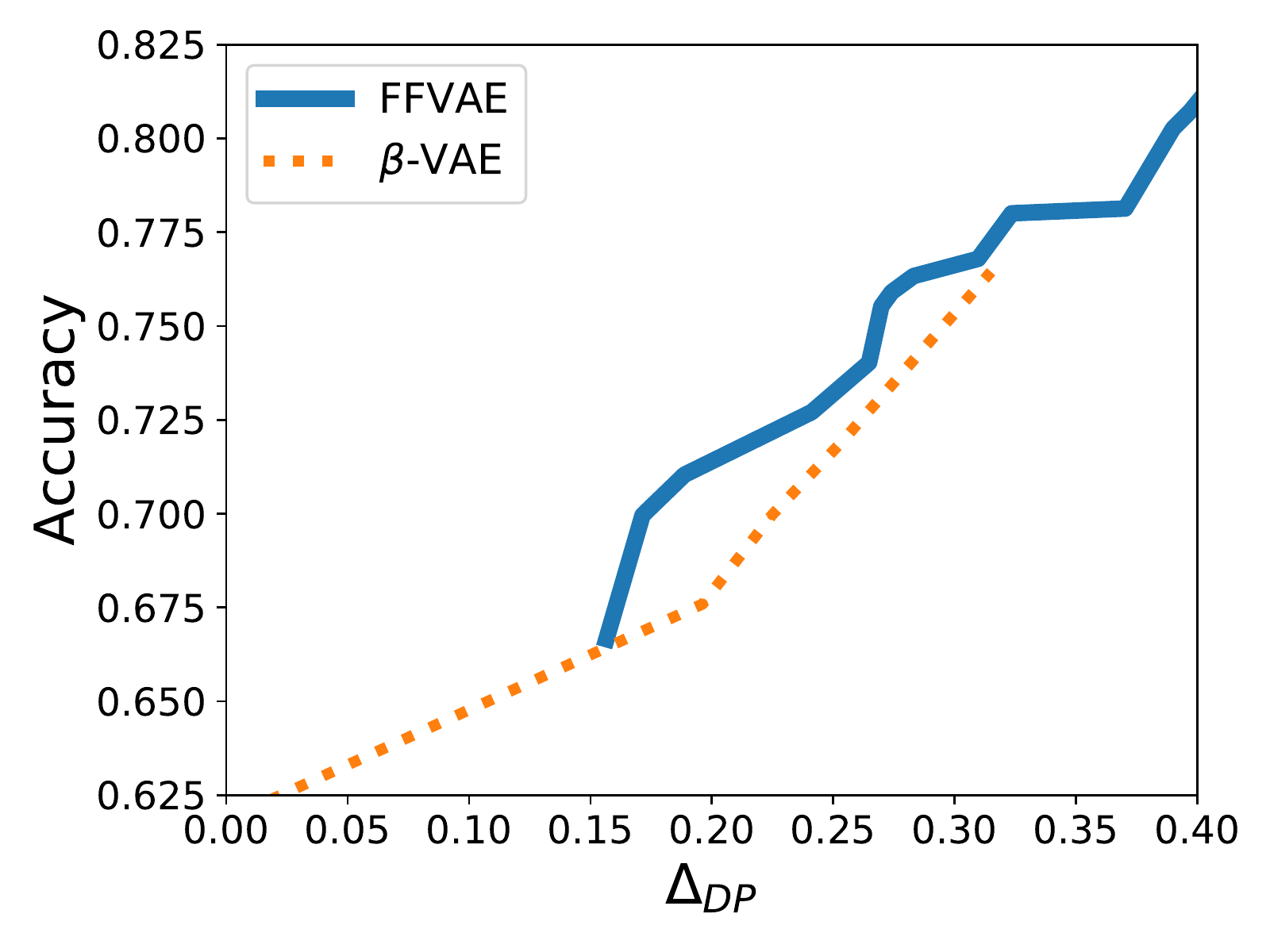}
\caption{$a$ = $\neg$ C $\wedge$ E}
\label{fig:predict_NOT-C-AND-E}
\end{subfigure}
\hfill
\begin{subfigure}[t]{\thirdFigWidth}
\includegraphics[width=\textwidth]{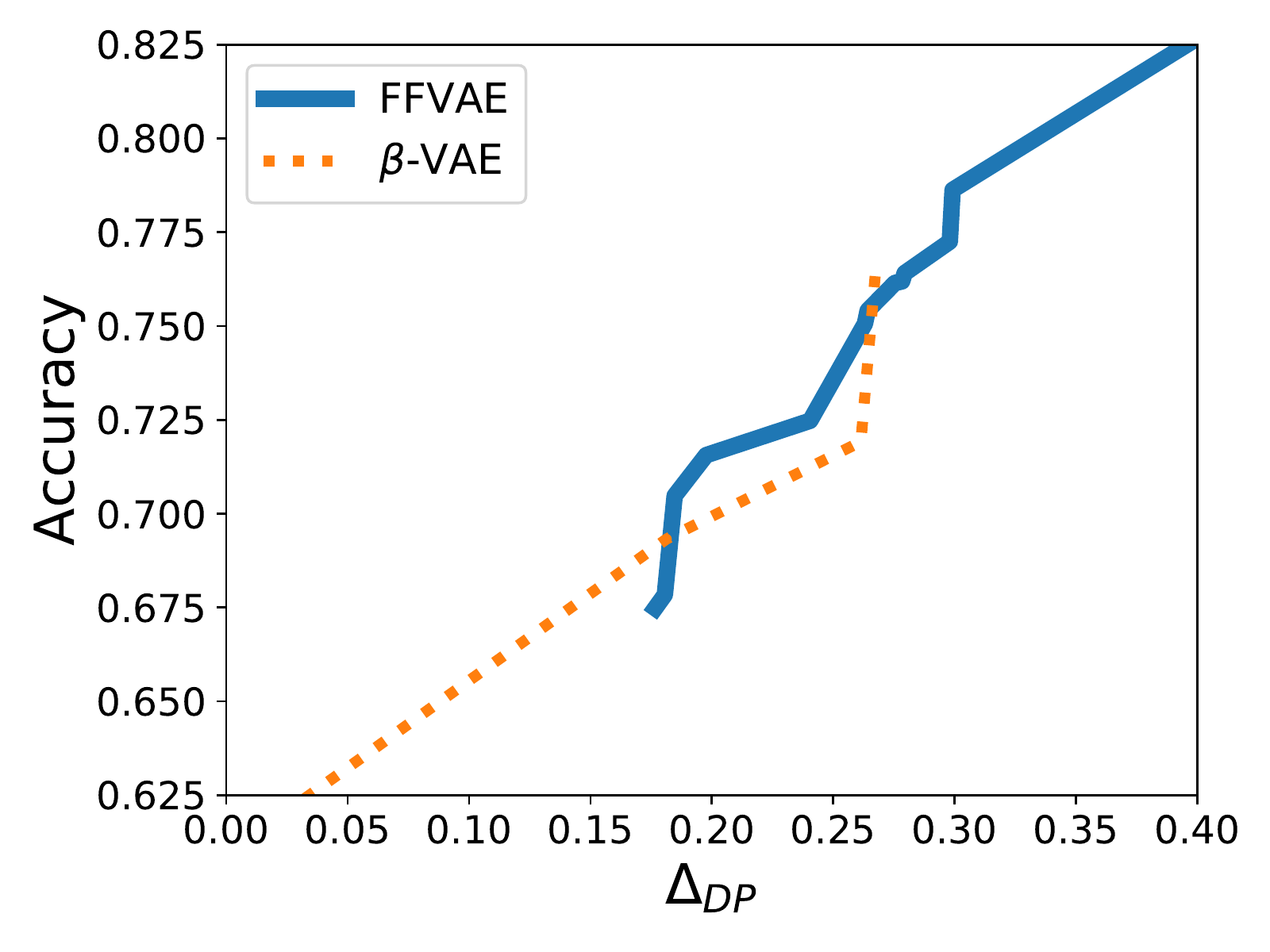}
\caption{$a$ = $\neg$ C $\wedge \neg$ E}
\label{fig:predict_NOT-C-AND-NOT-E}
\end{subfigure}
\hfill
\begin{subfigure}[t]{\thirdFigWidth}
\includegraphics[width=\textwidth]{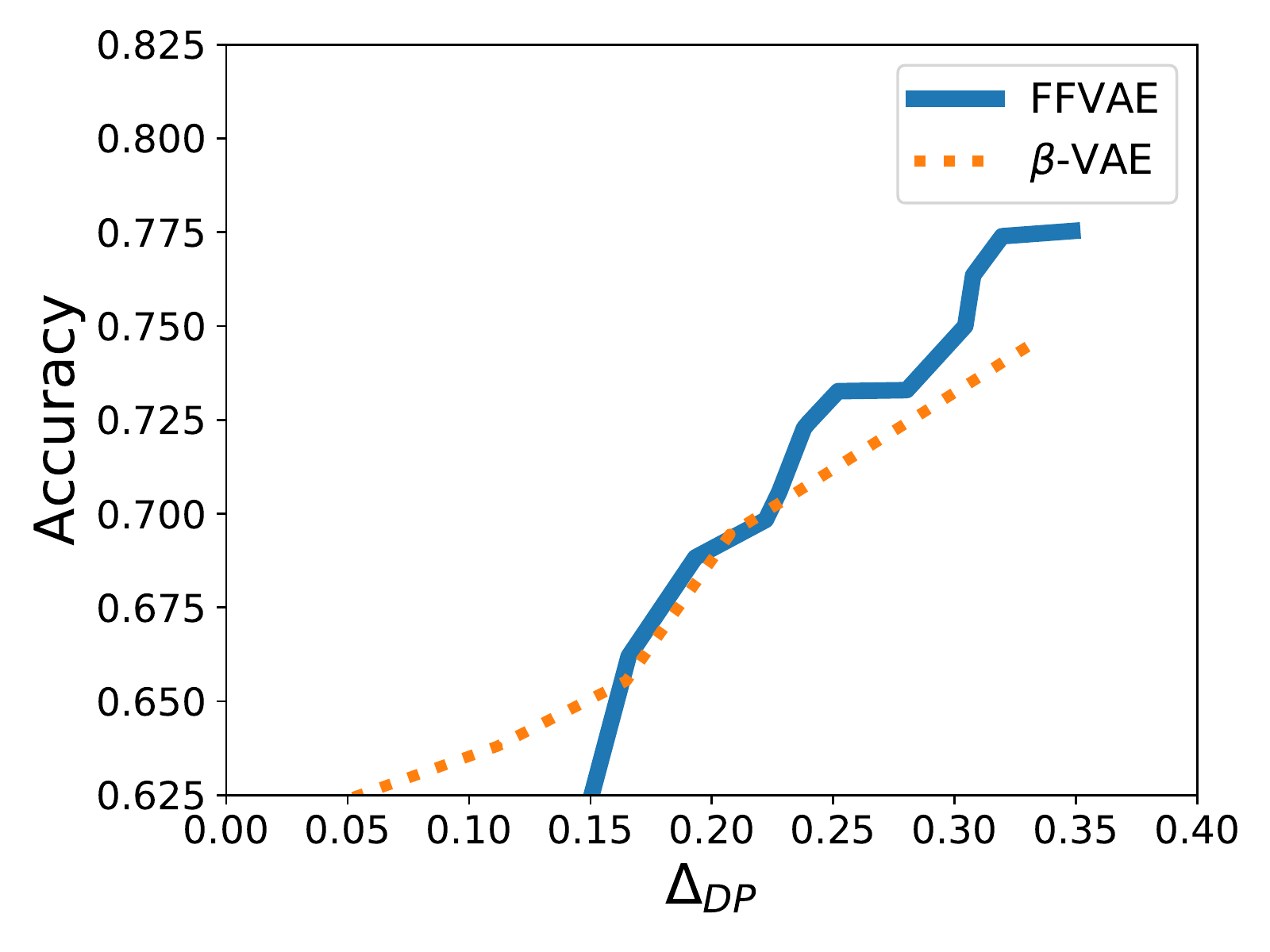}
\caption{$a$ = C $\wedge$ M}
\label{fig:predict_C-AND-M}
\end{subfigure}
\hfill
\begin{subfigure}[t]{\thirdFigWidth}
\includegraphics[width=\textwidth]{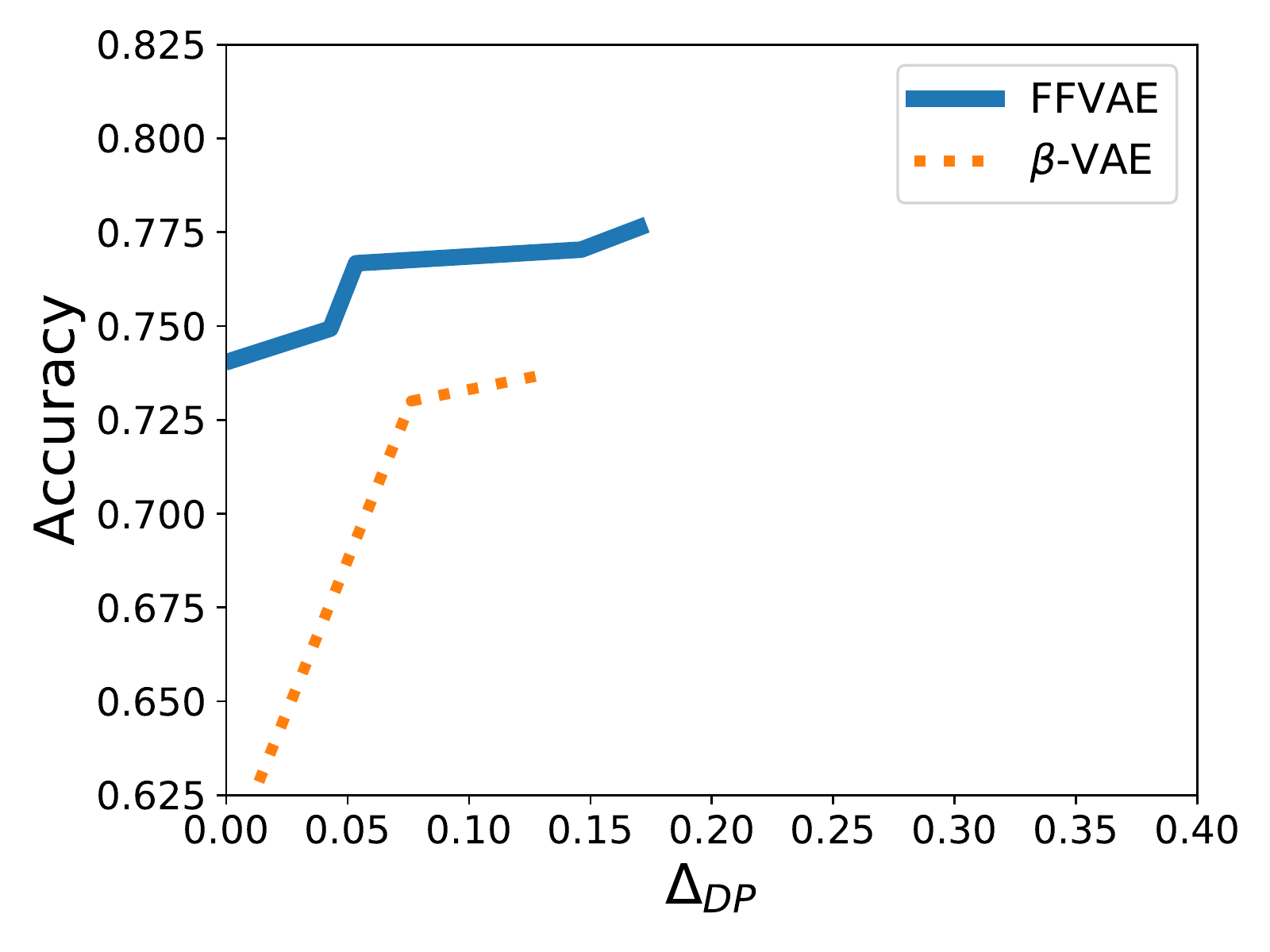}
\caption{$a$ = C $\wedge \neg$ M}
\label{fig:predict_C-AND-NOT-M}
\end{subfigure}
\hfill
\begin{subfigure}[t]{\thirdFigWidth}
\includegraphics[width=\textwidth]{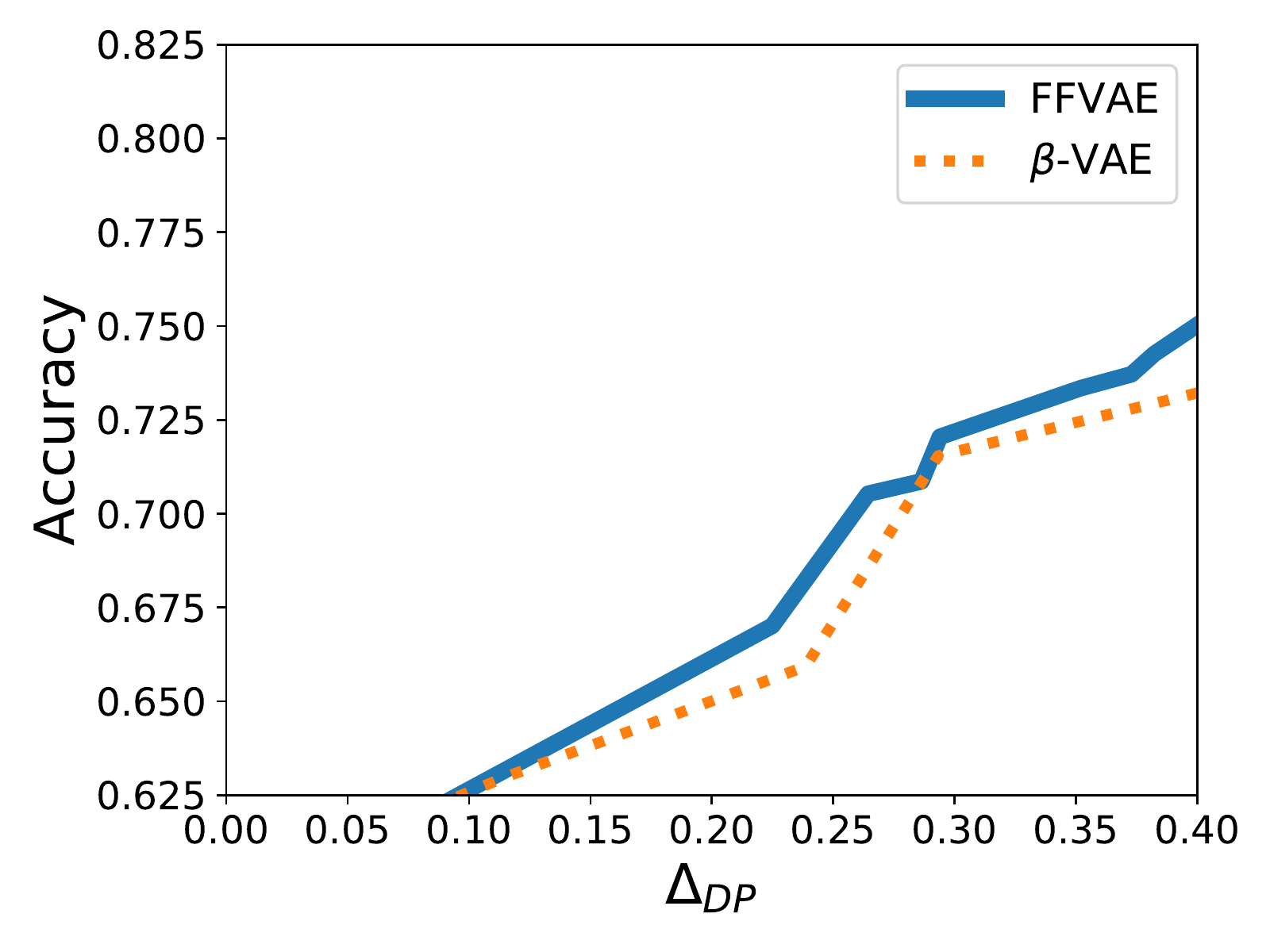}
\caption{$a$ = $\neg$ C $\wedge$ M}
\label{fig:predict_NOT-C-AND-M}
\end{subfigure}
\hfill
\begin{subfigure}[t]{\thirdFigWidth}
\includegraphics[width=\textwidth]{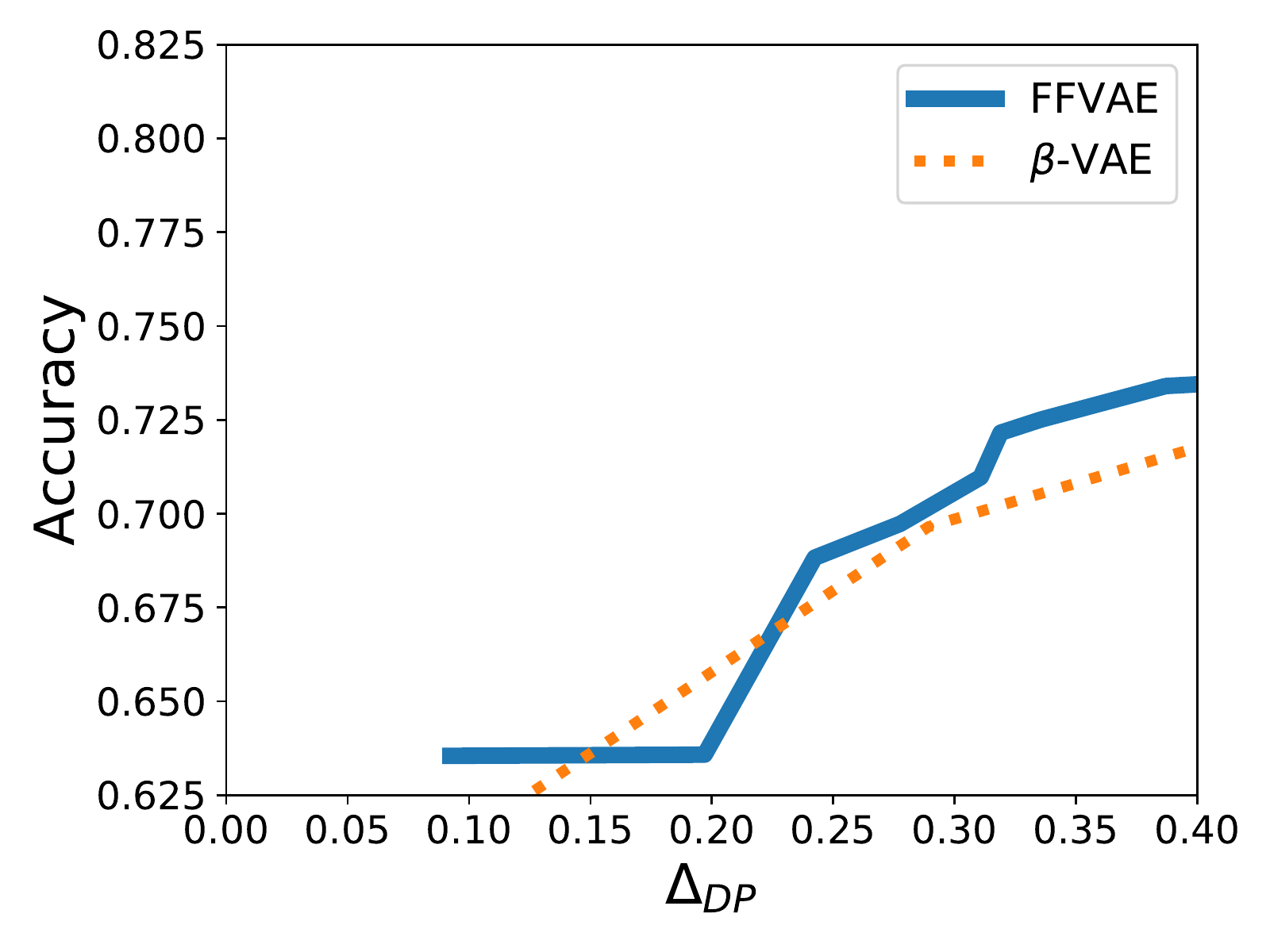}
\caption{$a$ = $\neg$ C $\wedge \neg$ M}
\label{fig:predict_NOT-C-AND-NOT-M}
\end{subfigure}
\hfill
\begin{subfigure}[t]{\thirdFigWidth}
\includegraphics[width=\textwidth]{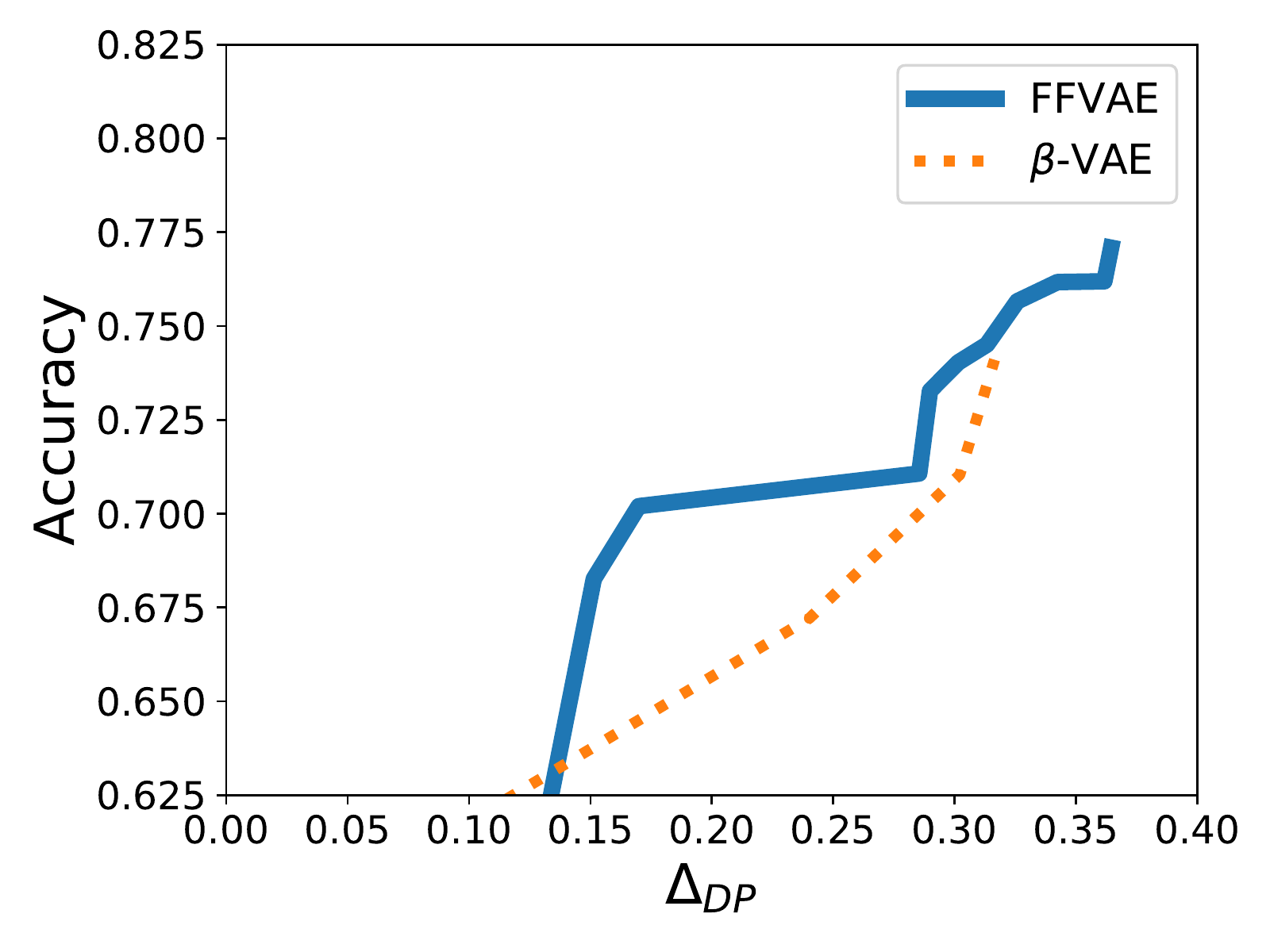}
\caption{$a$ = $\neg$ E $\wedge$ M}
\label{fig:predict_E-AND-M}
\end{subfigure}
\hfill

\begin{subfigure}[t]{\thirdFigWidth}
\includegraphics[width=\textwidth]{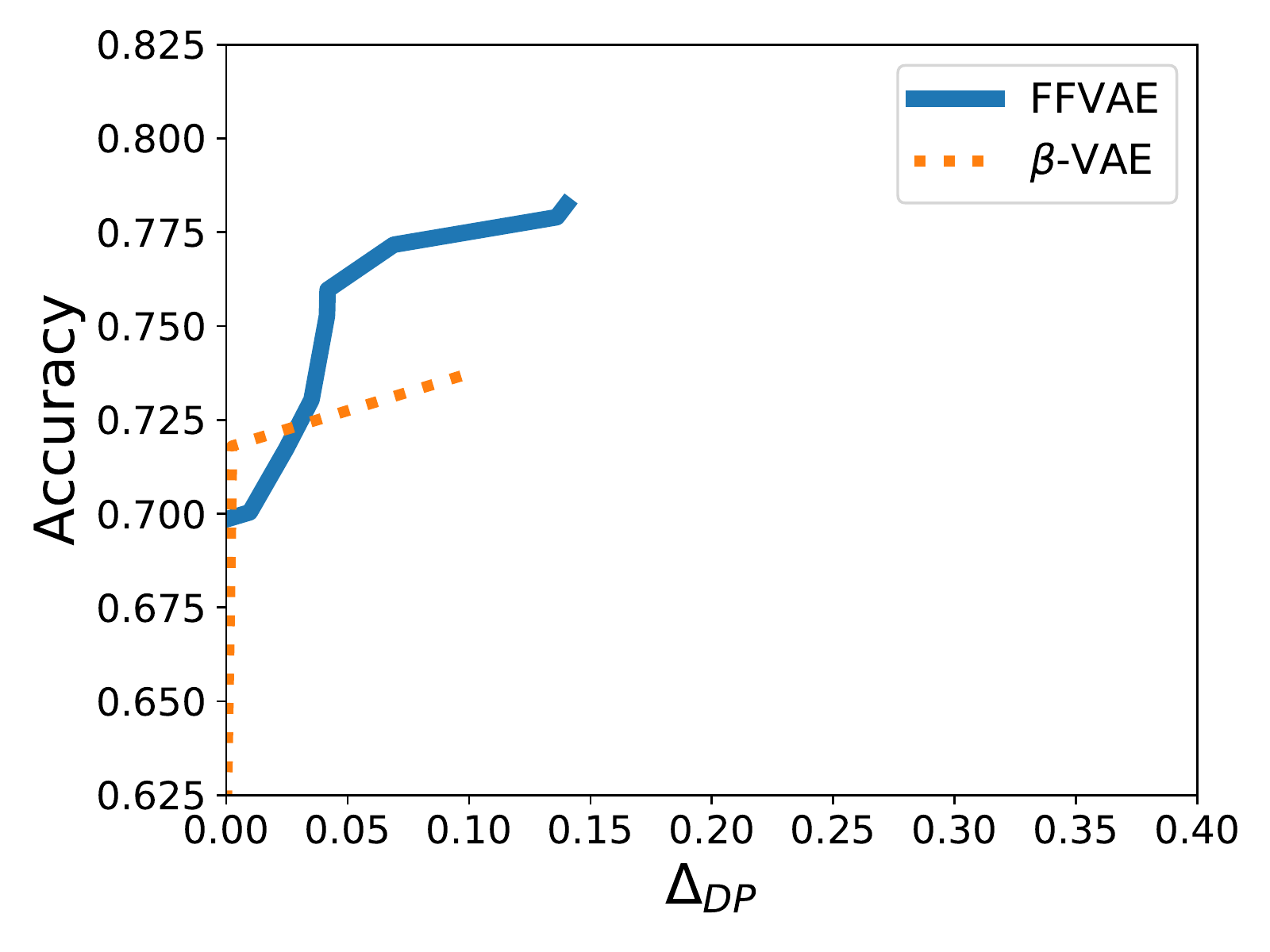}
\caption{$a$ = E $\wedge \neg$ M}
\label{fig:predict_E-AND-NOT-M}
\end{subfigure}
\hfill
\begin{subfigure}[t]{\thirdFigWidth}
\includegraphics[width=\textwidth]{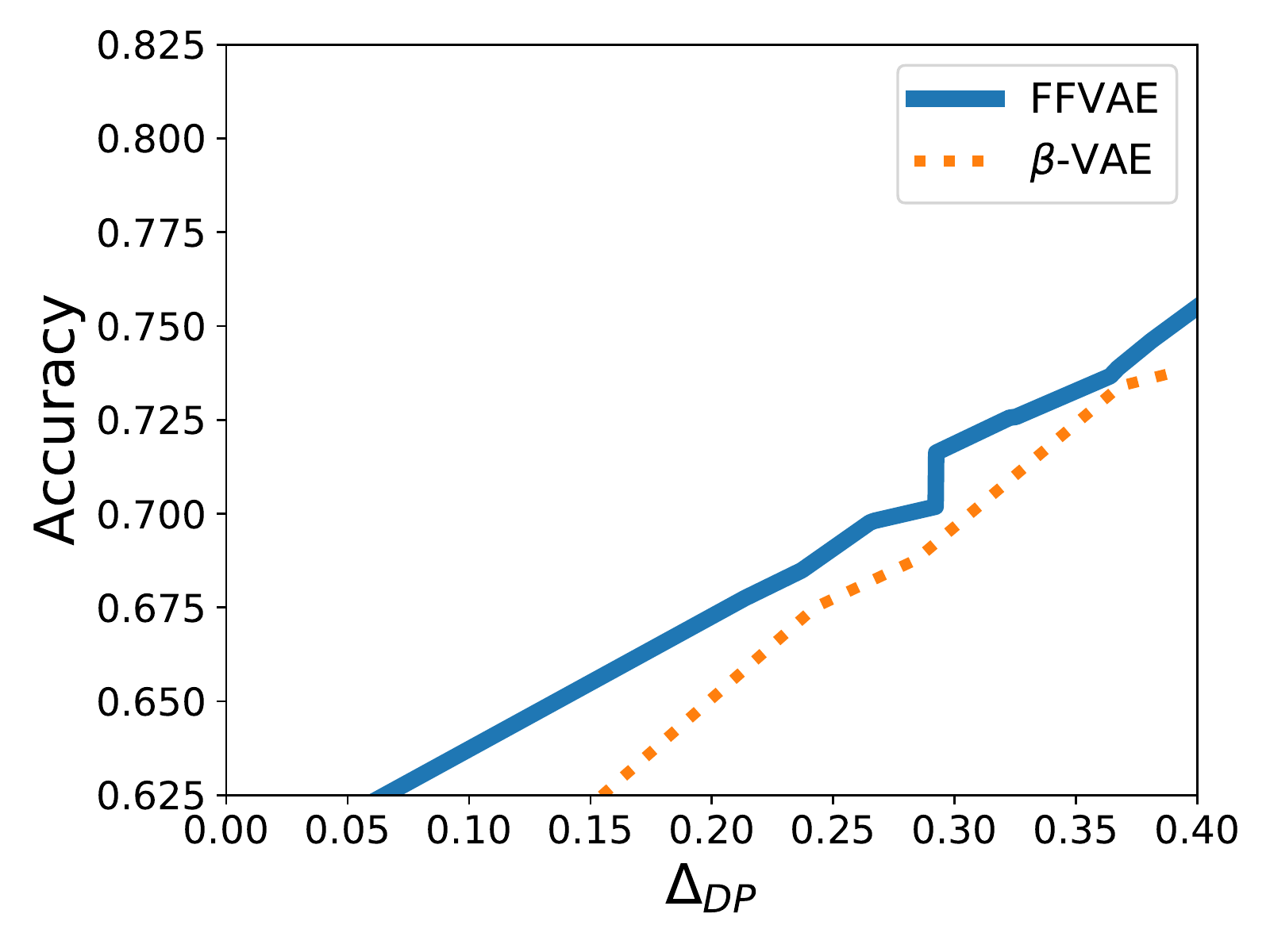}
\caption{$a$ = $\neg$ E $\wedge$ M}
\label{fig:predict_NOT-E-AND-M}
\end{subfigure}
\hfill
\begin{subfigure}[t]{\thirdFigWidth}
\includegraphics[width=\textwidth]{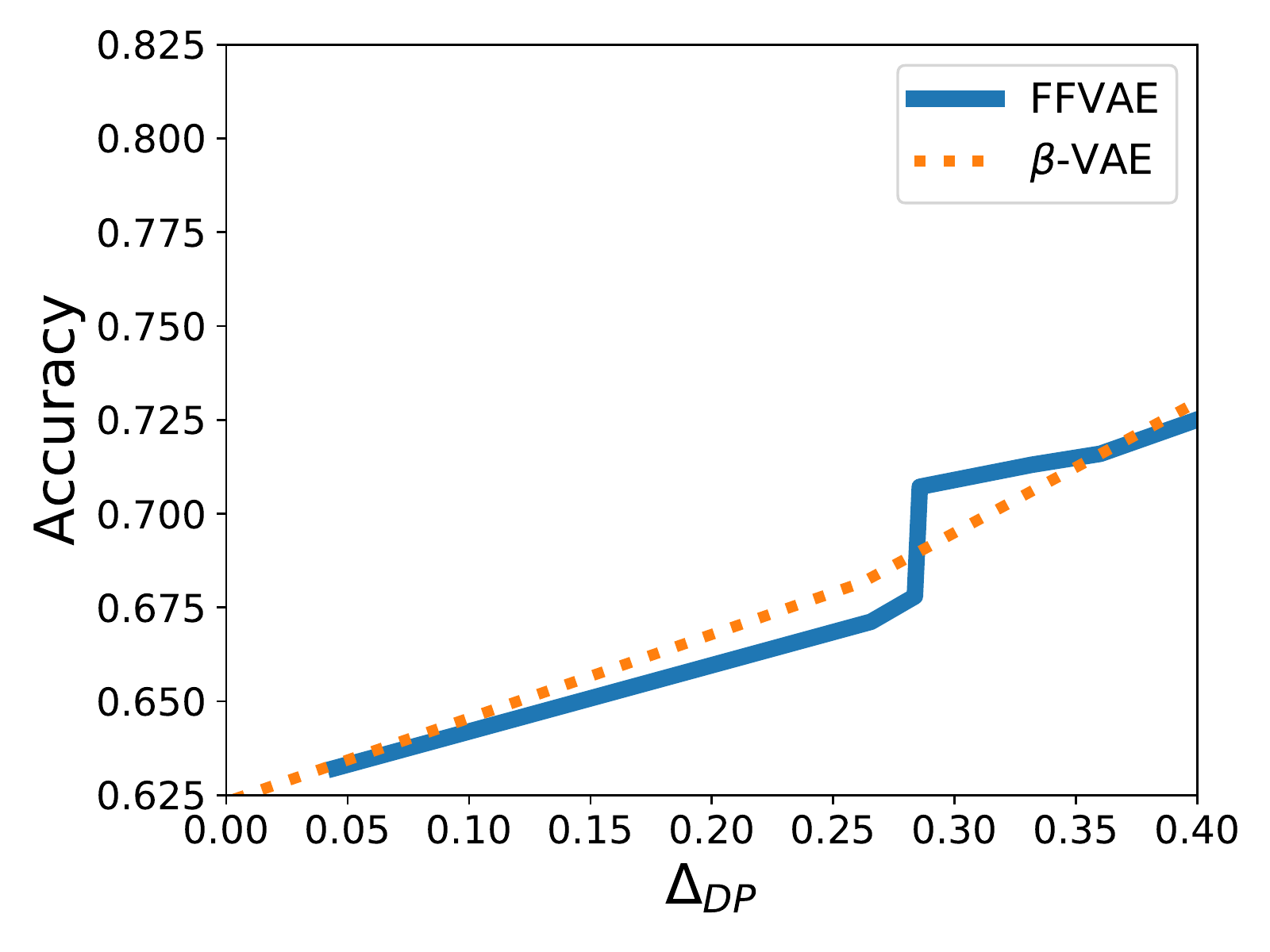}
\caption{$a$ = $\neg$ E $\wedge \neg$ M}
\label{fig:predict_NOT-E-AND-NOT-M}
\end{subfigure}
\hfill

\caption{
    Celeb-A subgroup fair classification results.
    Sensitive attributes: Chubby (C), Eyeglasses (E), and Male (M).
    $y$ = HeavyMakeup. 
    }
    \label{fig:celeba-pareto}
\end{figure}

\paragraph{Dataset}
The CelebA\footnote{\scriptsize \url{http://mmlab.ie.cuhk.edu.hk/projects/CelebA.html}} dataset contains over $200,000$ images of celebrity faces.
Each image is associated with 40 human-labeled binary attributes (OvalFace, HeavyMakeup, etc.).
We chose three attributes, Chubby, Eyeglasses, and Male as sensitive attributes\footnote{
We chose these attributes because they co-vary relatively weakly with each other (compared with other attribute triplets), but strongly with other attributes.
Nevertheless the rich correlation structure amongst all attributes makes this a challenging fairness dataset; it is difficult to achieve high accuracy and low $\Delta_{DP}$.
}, and report fair classification results on 3 groups and 12 two-attribute-conjunction subgroups only (for brevity we omit three-attribute conjunctions).
To our knowledge this is the first exploration of fair representation learning algorithms on the Celeb-A dataset.
As in the previous sections we train the encoders on the train set, then evaluate performance of MLP classifiers trained on the encoded test set.

\paragraph{Fair Classification}

We follow the fair classification audit procedure described above, where the held-out label HeavyMakeup---which was not used at encoder train time---is predicted by an MLP from the encoder representations.
When training the MLPs we take a fresh encoder sample for each minibatch (statically encoding the dataset with one encoder sample per image induced overfitting).
We found that training the MLPs on encoder means (rather than samples) increased accuracy but at the cost of very unfavorable $\Delta_{DP}$.
We also found that FactorVAE-style adversarial training does not scale well to this high-dimensional problem, so we instead optimize Equation \ref{eq:ffvae} using the biased estimator from \citet{chen2018isolating}.
Figure \ref{fig:celeba-pareto} shows Pareto fronts that capture the fairness-accuracy tradeoff for FFVAE and $\beta$-VAE.

While neither method dominates in this challenging setting, FFVAE achieves a favorable fairness-accuracy tradeoff across many of subgroups.
We believe that using sensitive attributes as side information gives FFVAE an advantage over $\beta$-VAE in predicting the held-out label.
In some cases (e.g., $a$=$\neg$E$\wedge$M) FFVAE achieves better accuracy at all $\Delta_{DP}$ levels, while in others (e.g., $a$=$\neg$C$\wedge\neg$E)
, FFVAE did not find a low-$\Delta_{DP}$ solution.
We believe Celeb-A--with its many high dimensional data and rich label correlations---is a useful test bed for subgroup fair machine learning algorithms, and we are encouraged by the reasonably robust performance of FFVAE in our experiments.

\section{Discussion}
In this paper we discussed how disentangled representation learning aligns with the goals of subgroup fair machine learning, and presented a method for learning a structured latent code using multiple sensitive attributes.
The proposed model, FFVAE, provides \textit{flexibly fair} representations, which can be modified simply and compositionally at test time to yield a fair representation with respect to multiple sensitive attributes and their conjunctions, even when test-time sensitive attribute labels are unavailable.
Empirically we found that FFVAE disentangled sensitive sources of variation in synthetic image data, even in the challenging scenario where attributes and labels correlate.
Our method compared favorably with baseline disentanglement algorithms on downstream fair classifications by achieving better parity for a given accuracy budget across several group and subgroup definitions.
FFVAE also performed well on the Communities \& Crime and Celeb-A dataset, although none of the models performed robustly across all possible subgroups in the real-data setting.
This result reflects the difficulty of subgroup fair representation learning and motivates further work on this topic.

There are two main directions of interest for future work.
First is the question of fairness metrics: a wide range of fairness metrics beyond demographic parity have been proposed \citep{hardt2016equality,pleiss2017fairness}.
Understanding how to learn flexibly fair representations with respect to other metrics is an important step in extending our approach.
Secondly, robustness to distributional shift presents an important challenge in the context of both disentanglement and fairness.
In disentanglement, we aim to learn independent factors of variation.
Most empirical work on evaluating disentanglement has used synthetic data with uniformly distributed factors of variation, but this setting is unrealistic. 
Meanwhile, in fairness, we hope to learn from potentially biased data distributions, which may suffer from both undersampling and systemic historical discrimination.
We might wish to imagine hypothetical ``unbiased'' data or compute robustly fair representations, but must do so given the data at hand.
While learning fair or disentangled representations from real data remains a challenge in practice, we hope that this investigation serves as a first step towards understanding and leveraging the relationship between the two areas.

\bibliography{refs}
\bibliographystyle{icml2019}

\clearpage
\appendix
\section{Discriminator approximation of total correlation} \label{sec:disc_approx}
This section describes how density ratio estimation is implemented to train the FFVAE encoder.
We follow the approach of \citet{kim2018disentangling}.
\paragraph{Generating Samples}
The binary classifier adversary seeks to discriminate between
\begin{itemize}
    \item $[z,b] \sim q(z,b)$, ``true'' samples from the aggregate posterior; and
    \item $[z',b'] \sim q(z)\prod_j q(b_j)$, ``fake'' samples from the product of the marginal over $z$ and the marginals over each $b_j$.
\end{itemize}
At train time, after splitting the latent code $[z^i,b^i]$ of the $i$-example along the dimensions of $b$ as $[z^i, b_0^i ... b_j^i]$, the minibatch index order for each subspace is then randomized, simulating samples from the product of the marginals; these dimension-shuffled samples retain the same marginal statistics as ``real'' (unshuffled) samples, but with joint statistics between the subspaces broken.
The overall minibatch of encoder outputs contains twice as many examples as the original image minibatch, and comprises equal number of ``real'' and ``fake'' samples.

As we describe below, the encoder output minibatch is used as training data for the adversary, and the error is backpropagated to the encoder weights so the encoder can better fool the adversary.
 If a strong adversary can do no better than random chance, then the desired independence property has been achieved.

\paragraph{Discriminator Approximation}
Here we summarize the the approximation of the $D_{KL}(q(z,b)||q(z)\prod_{j} q(b_j))$ term from equation \ref{eq:ffvae}.
Let $u \in \{0, 1\}$ be an indicator variable with $u=1$ indicating $[z,b] \sim q(z,b)$ comes from a minibatch of ``real'' encoder distributions, while $u=1$ indicating $[z',b'] \sim q(z)\prod_j(b_j)$ is drawn from a ``fake'' minibatch of shuffled samples, i.e., is drawn from the product of the marginals of the aggregate posterior.
The discriminator network outputs the probability that vector $[z,b]$ is a ``real'' sample, i.e., 
$d(u|z,b)=\text{Bernoulli}(u|\sigma(\theta_{d}(z,b))$ 
where $\theta_d(z,b)$ 
is the discriminator and $\sigma$ is the sigmoid function.
If the discriminator is well-trained to distinguish between ``real'' and ``fake'' samples then we have 
\begin{align}
    \log d(u=1|z,b) - \log d(u=0|z,b) &\approx \nonumber \\
    \log q(z,b) - \log q(z)\prod_j q(b_j).
\end{align}
We can substitute this into the KL divergence as 
\begin{align}
    &D_{KL}(q(z,b)||q(z)\prod_{j} q(b_j)) = \nonumber \\
    &\quad\quad \E_{q(z,b)}[ \log q(z,b) - \log q(z)\prod_j q(b_j) ] \approx \nonumber \\
    &\quad\quad \E_{q(z,b)}[ \log d(u=1|z,b) - \log d(u=0|z,b)].
\end{align}

Meanwhile the discriminator is trained by minimizing the standard cross entropy loss
\begin{align}
    L_{\text{Disc}}(d) &= \E_{z,b\sim q(z,b)} [\log d(u=1|z,b)] \nonumber \\
    &\quad\quad + \E_{z',b'\sim q(z)\prod_j q(b_j)} [\log ( 1 - d(u=0|z',b') )],
\end{align}
w.r.t. the parameters of $d(u|z,b)$.
This ensures that the discriminator output $\theta_d(z,b)$ is a calibrated approximation of the log density $\log\frac{q(z,b)}{q(z)\prod_j q(b_j)}$.

$L_{\text{Disc}}(d)$ and $L_{\text{FFVAE}}(p,q)$ (Equation \ref{eq:ffvae}) are then optimized in a min-max fashion.
In our experiments we found that single-step alternating updates using optimizers with the same settings sufficed for stable optimization.

\section{DSpritesUnfair Generation} \label{sec:dsprites-generation}
The original DSprites dataset has six ground truth factors of variation (FOV): 
\begin{itemize}
    \item Color: white
    \item Shape: square, ellipse, heart
    \item Scale: 6 values linearly spaced in $[0.5, 1]$
    \item Orientation: 40 values in $[0, 2\pi]$
    \item XPosition: 32 values in $[0, 1]$
    \item YPosition: 32 values in $[0, 1]$
\end{itemize}

In the original dataset the joint distribution over all FOV factorized; each FOV was considered independent.
In our dataset, we instead sample such that the FOVs Shape and X-position correlate.
We associate an index with each possible value of each attribute, and then sample a (Shape, X-position) pair with probability proportional to $(\frac{i_S}{n_S})^{q_S} + (\frac{i_X}{n_X})^{q_X}$, where $i, n, q$ are the indices, total number of options, and a real number for each of Shape and X-position ($S, X$ respectively).
We use $q_S = 1, q_X = 3$. 
All other attributes are sampled uniformly, as in the standard version of DSprites.

We binarized the factors of variation by using the boolean outputs of the following operations:
\begin{itemize}
    \item Color $\geq 1$
    \item Shape $\geq 1$
    \item Scale $\geq 3$
    \item Rotation $\geq 20$
    \item XPosition $\geq 16$
    \item YPosition $\geq 16$
\end{itemize}

\section{DSprites Architectures}\label{sec:archs}
The architectures for the convolutional encoder $q(z,b|x)$, decoder $q(x|z,b)$, and FFVAE discriminator are specified as follows.
{\tiny

\begin{verbatim}
import torch
from torch import nn

class Resize(torch.nn.Module):
    def __init__(self, size):
        super(Resize, self).__init__()
        self.size = size

    def forward(self, tensor):
        return tensor.view(self.size)

class ConvEncoder(nn.Module):
    def __init__(self, im_shape=[64, 64], latent_dim=10, n_chan=1):
        super(ConvEncoder, self).__init__()

        self.f = nn.Sequential(
            Resize((-1,n_chan,im_shape[0],im_shape[1])),
            nn.Conv2d(n_chan, 32, 4, 2, 1),
            nn.ReLU(True),
            nn.Conv2d(32, 32, 4, 2, 1),
            nn.ReLU(True),
            nn.Conv2d(32, 64, 4, 2, 1),
            nn.ReLU(True),
            nn.Conv2d(64, 64, 4, 2, 1),
            nn.ReLU(True),
            Resize((-1,1024)),
            nn.Linear(1024, 128),
            nn.ReLU(True),
            nn.Linear(128, 2*latent_dim)
            )
        self.im_shape = im_shape
        self.latent_dim = latent_dim

    def forward(self, x):
        mu_and_logvar = self.f(x)
        mu = mu_and_logvar[:, :self.latent_dim]
        logvar = mu_and_logvar[:, self.latent_dim:]
        return mu, logvar

class ConvDecoder(nn.Module):
    def __init__(self, im_shape=[64, 64], latent_dim=10, n_chan=1):
        super(ConvDecoder, self).__init__()

        self.g = nn.Sequential(
            nn.Linear(latent_dim, 128),
            nn.ReLU(True),
            nn.Linear(128, 1024),
            nn.ReLU(True),
            Resize((-1,64,4,4)),
            nn.ConvTranspose2d(64, 64, 4, 2, 1),
            nn.ReLU(True),
            nn.ConvTranspose2d(64, 32, 4, 2, 1),
            nn.ReLU(True),
            nn.ConvTranspose2d(32, 32, 4, 2, 1),
            nn.ReLU(True),
            nn.ConvTranspose2d(32, n_chan, 4, 2, 1),
            )

    def forward(self, z):
        x = self.g(z)
        return x.squeeze()

class Discriminator(nn.Module):
    def __init__(self, n):
        super(Discriminator, self).__init__()
        self.model = nn.Sequential(
            nn.Linear(n, 1000),
            nn.LeakyReLU(0.2, inplace=True),
            nn.Linear(1000, 1000),
            nn.LeakyReLU(0.2, inplace=True),
            nn.Linear(1000, 1000),
            nn.LeakyReLU(0.2, inplace=True),
            nn.Linear(1000, 1000),
            nn.LeakyReLU(0.2, inplace=True),
            nn.Linear(1000, 1000),
            nn.LeakyReLU(0.2, inplace=True),
            nn.Linear(1000, 2),
            )
        selftmax = nn.Softmax(dim=1)

    def forward(self, zb):
        logits = self.model(zb)
        probs = nn.Softmax(dim=1)(logits)
        return logits, probs
\end{verbatim}
}

\section{DSpritesUnfair Training Details}\label{sec:dsprites-training-details}
All network parameters were optimized using the Adam \citep{kingma2014adam}, with learning rate 0.001.
Architectures are specified in Appendix \ref{sec:archs}.
Our encoders trained $3 \times 10^5$ iterations with minibatch size 64 (as in \citet{kim2018disentangling}).
Our MLP classifier has two hidden layers with 128 units each, and is trained with patience of 5 epochs on validation loss.

\section{Mutual Information Gap} \label{sec:mig}
\begin{figure}[ht!]
%
\begin{subfigure}[t]{0.23\textwidth}
\includegraphics[width=\textwidth]{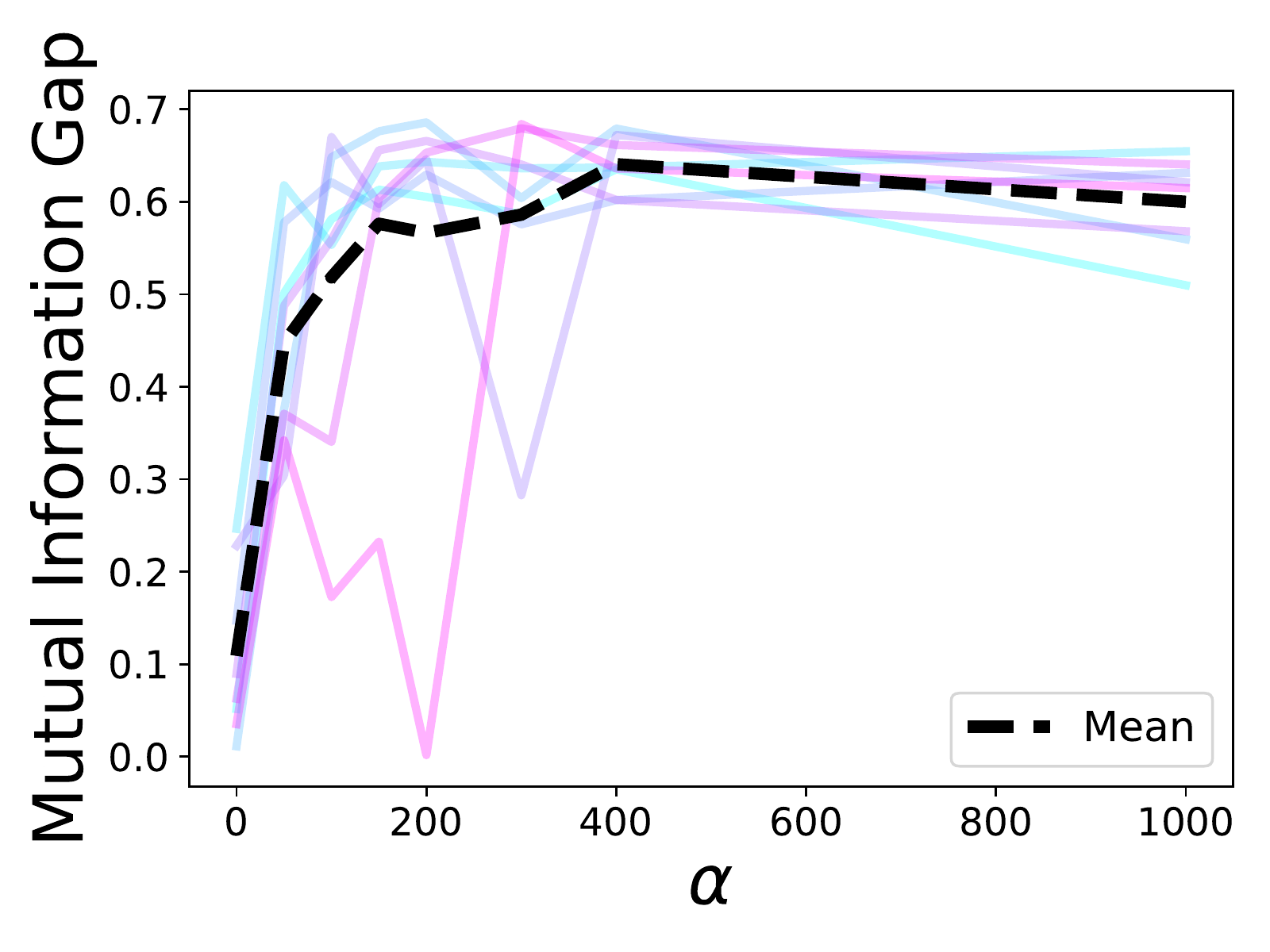}
\caption{Color is $\gamma$, brighter colours $\longrightarrow$ higher values}
\label{fig:dsprites-mig-lines}
\end{subfigure}
\hfill
\begin{subfigure}[t]{0.23\textwidth}
\includegraphics[width=\textwidth]{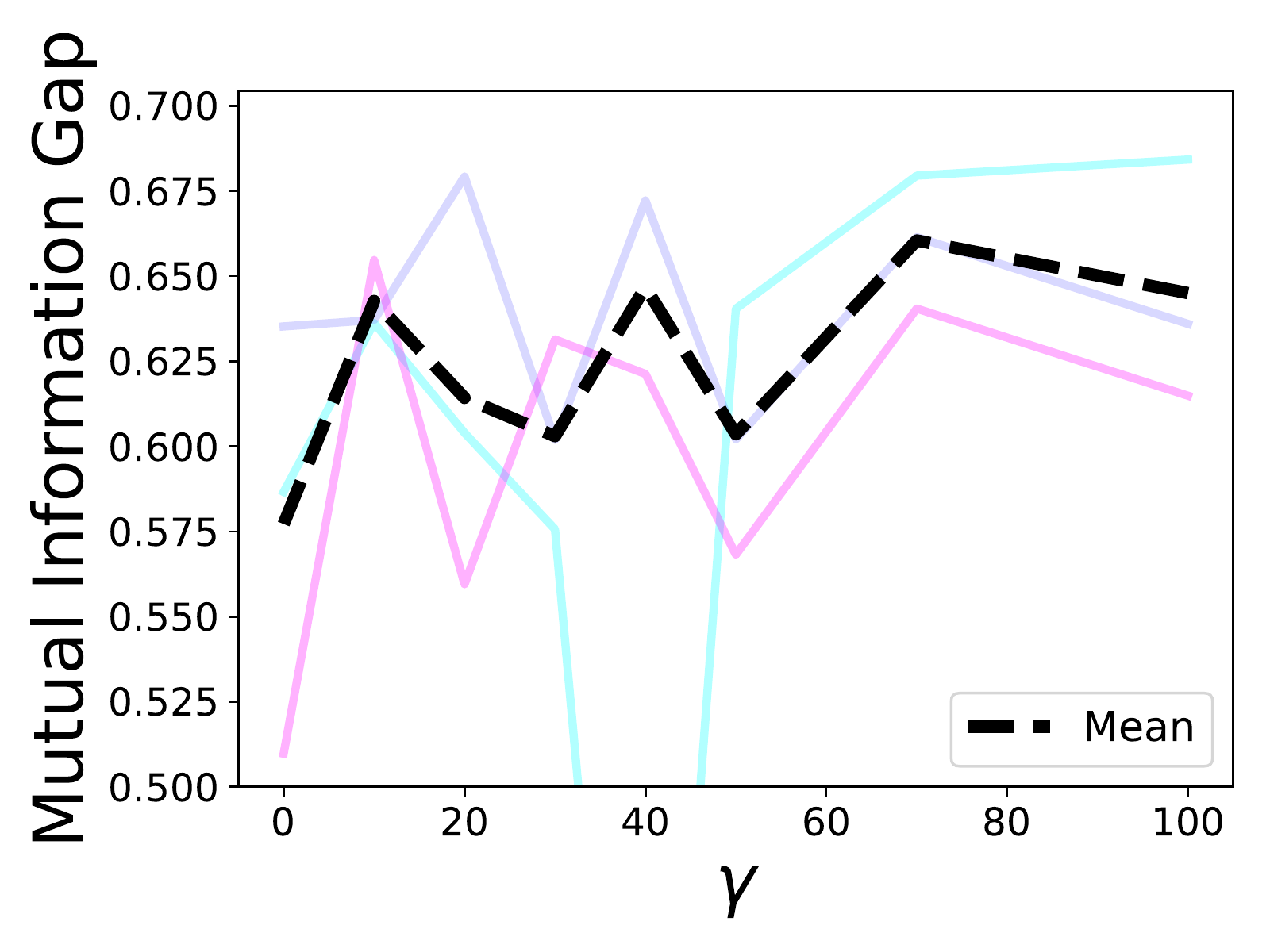}
\caption{Colour is $\alpha$, brighter colors $\longrightarrow$ higher values}
\label{fig:dsprites-mig-lines-high-alpha}
\end{subfigure}

\caption{
    Mutual Information Gap (MIG) for various $(\alpha, \gamma)$ settings of the FFVAE. 
    In Fig. \ref{fig:dsprites-mig-lines}, each line is a different value of
    $\gamma \in [10, 20, 30, 40, 50, 70, 100]$, with brighter colors indicating larger values of $\gamma$.
    In Fig. \ref{fig:dsprites-mig-lines-high-alpha}, each line is a different value of
    $\alpha \in [300, 400, 1000]$, with brighter colors indicating larger values of $\alpha$.
    Models trained on DspritesUnfair, MIG calculated on Dsprites.
    Higher MIG is better.
    Black dashed line indicates mean (with outliers excluded).
    $\alpha = 0$ is equivalent to the FactorVAE.
    }
    \label{fig:dsprites-mig}
\end{figure}

\paragraph{Evaluation Criteria}
Here we analyze the encoder mutual information in the synthetic setting of the DSpritesUnfair dataset, where we know the ground truth factors of variation.
In Fig. \ref{fig:dsprites-mig}, we calculate the \textit{Mutual Information Gap (MIG)} \citep{chen2018isolating} of FFVAE across various hyperparameter settings.
With $J$ latent variables $z_j$ and $K$ factors of variation $v_k$, MIG is defined as 
\begin{equation}
    \frac{1}{K} \sum_{k=1}^{K} \frac{1}{H(v_k)} (MI (z_{j_k}; v_k) - \max_{j \neq j_k} MI (z_j; v_k))
\end{equation}
where $j_k = \underset{j}{\argmax} MI (z_j; v_k)$, $MI(\cdot; \cdot)$ denotes mutual information, and $K$ is the number of factors of variation.
Note that we can only compute this metric in the synthetic setting where the ground truth factors of variation are known.
MIG measures the difference between the latent variables which have the highest and second-highest $MI$ with each factor of variation, rewarding models which allocate one latent variable to each factor of variation.
We test our disentanglement by training our models on a biased version of DSprites, and testing on a balanced version (similar to the ``skewed'' data in \citet{chen2018isolating}).
This allows us to separate out two sources of correlation --- the correlation existing across the data, and the correlation in the model's learned representation.

\paragraph{Results}

In Fig. \ref{fig:dsprites-mig-lines}, we show that MIG increases with $\alpha$, providing more evidence that the supervised structure of the FFVAE can create disentanglement.
This improvement holds across values of $\gamma$, except for some training instability for the highest values of $\gamma$.
It is harder to assess the relationship between $\gamma$ and MIG, due to increased instability in training when $\gamma$ is large and $\alpha$ is small.
However, in Fig. \ref{fig:dsprites-mig-lines-high-alpha}, we look only at $\alpha \geq 300$, and note that in this range, MIG improves as $\gamma$ increases.
See Appendix \ref{sec:mig} for more details.

\begin{figure}[ht!]
%
\begin{subfigure}[t]{0.23\textwidth}
\includegraphics[width=\textwidth]{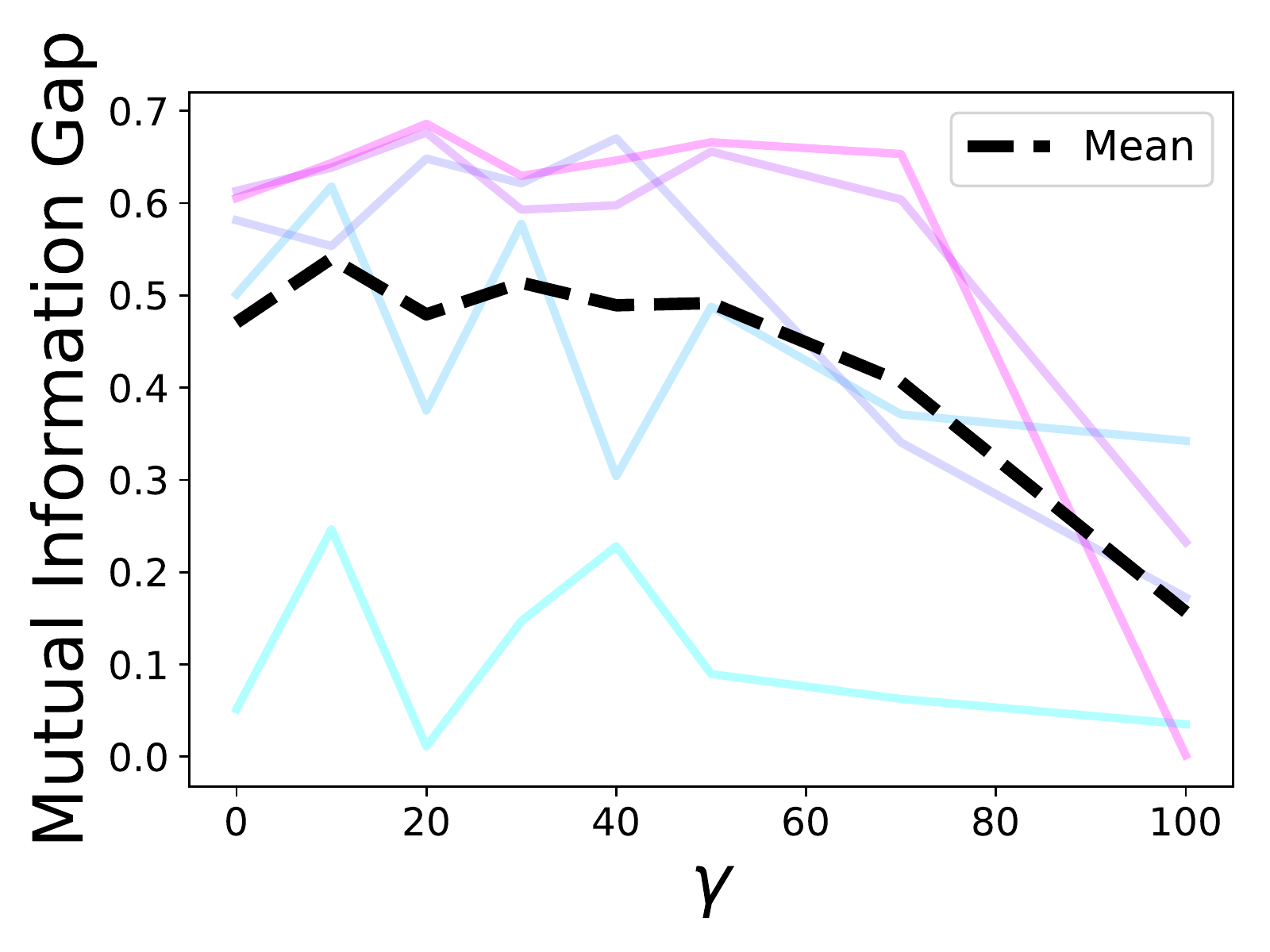}
\caption{Colour is $\alpha$}
\label{fig:dsprites-mig-lines-low-alpha}
\end{subfigure}
\hfill
\begin{subfigure}[t]{0.23\textwidth}
\includegraphics[width=\textwidth]{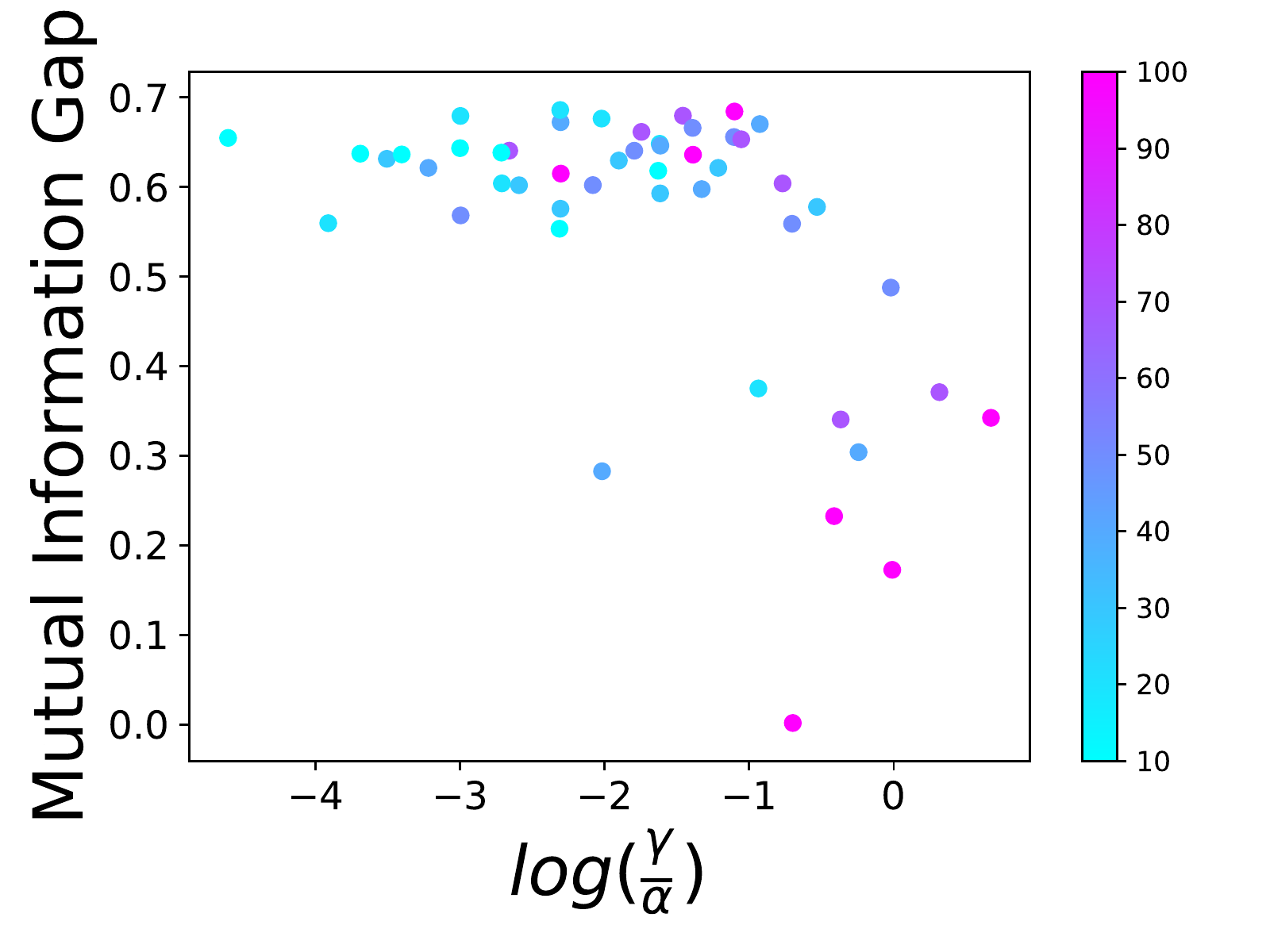}
\caption{Colour is $\gamma$}
\label{fig:dsprites-mig-ratio}
\end{subfigure}

\caption{
    Mutual Information Gap (MIG) for various $(\alpha, \gamma)$ settings of the FFVAE. 
    In Fig. \ref{fig:dsprites-mig-lines-low-alpha}, each line is a different value of
    $\alpha \in [0, 50, 100, 150, 200]$, with brighter colours indicating larger values of $\alpha$.
    In Fig. \ref{fig:dsprites-mig-ratio}, all combinations with $\alpha, \gamma > 0$ are shown.
    Models trained on DspritesUnfair, MIG calculated on Dsprites.
    Higher MIG is better.
    Black dashed line indicates mean (outliers excluded).
    $\alpha = 0$ is equivalent to the FactorVAE.
    }
    \label{fig:dsprites-mig-appendix}
\end{figure}

In Fig. \ref{fig:dsprites-mig-lines-low-alpha}, we show that for low values of $\alpha$, increasing $\gamma$ leads to worse MIG, likely due to increased training instability.
This is in contrast to Fig. \ref{fig:dsprites-mig-lines-high-alpha}, which suggests that for high enough $\alpha$, increasing $\gamma$ can improve MIG.
This leads us to believe that $\alpha$ and $\gamma$ have a complex relationship with respect to disentanglement and MIG.

To better understand the relationship between these two hyperparameters, we examine how MIG varies with the ratio $\frac{\gamma}{\alpha}$ in Fig. \ref{fig:dsprites-mig-ratio}.
In We find that in general, a higher ratio yields lower MIG, but that the highest MIGs are around $\log\frac{\gamma}{\alpha} = -2$, with a slight tailing off for smaller ratios.
This indicates there is a dependent relationship between the values of $\gamma$ and $\alpha$.

\paragraph{Discussion}
What does it mean for our model to demonstrate disentanglement on test data drawn from a new distribution?
For interpretation, we can look to the causal inference literature, where one goal is to produce models that are robust to certain interventions in the data generating process \citep{rothlenhauser2018anchor}.
We can interpret Figure \ref{fig:dsprites-mig} as evidence that our learned representations are (at least partially) invariant to \textit{interventions} on $a$.
This property relates to counterfactual fairness, which requires that models be robust with respect to counterfactuals along $a$ \citep{kusner2017counterfactual}.

\end{document}